\documentclass{article} \usepackage{paper_style,times}

\usepackage{amsmath,amsfonts,bm}

\newcommand{\ourmech}{PAT}

\def\eqref#1{equation~\ref{#1}}

\def\1{\bm{1}}

\DeclareMathAlphabet{\mathsfit}{\encodingdefault}{\sfdefault}{m}{sl}
\SetMathAlphabet{\mathsfit}{bold}{\encodingdefault}{\sfdefault}{bx}{n}

\usepackage{hyperref}
\usepackage{url}
\usepackage{booktabs}
\usepackage{graphicx}
\usepackage{capt-of}
\usepackage{tabularx,array,booktabs}
\usepackage{multirow}
\usepackage[table]{xcolor}
\usepackage{graphicx}
\usepackage[export]{adjustbox}
\usepackage[dvipsnames]{xcolor}

\title{Personalize at Test Time: Learning User Preferences for Image Generation}

\author{Jiamu Bai$^\dagger$ \thanks{Work done during Jiamu Bai's internship at Meta. $\dagger$: co-first author.}  \\
Penn State University\\
\And
Jiaming Hu$^\dagger$ \\
Boston University \\
\And
Yanhong Wu \\
Meta Inc. \\
\And
Zellux Wang\\
Meta Inc. 
}

\paperfinalcopy 
\begin{document}

\maketitle
\begin{abstract}

Diffusion models can generate high-quality images, yet aligning their outputs with individual user preferences remains challenging. A key bottleneck is accurately modeling diverse user preferences from limited feedback. Existing approaches often rely on labor-intensive manual preference annotations or vision-language models(VLM) to extract preference information from user interaction histories, introducing substantial annotation or computational costs that limit scalability. We propose an approach that learns personalized reward models directly from users’ historical image preference pairs. First, we use an autoencoder to compress hundreds of visual attributes into 50 attribute-anchored preference dimensions and train an evaluator to score images along these dimensions. We then represent each user’s preferences as a linear combination of the shared dimension scores, estimating the user-specific weights by maximizing the likelihood of their observed pairwise preferences under the Bradley–Terry model. This formulation reduces per-user adaptation to optimizing a low-dimensional weight vector, simplifying optimization and enabling data-efficient personalization from sparse feedback. The learned personalized rewards guide image generation at inference time while keeping the diffusion model frozen. Experiments on real-user preference data show that our approach achieves approximately 77\% held-out pairwise preference prediction accuracy and improves the alignment of generated images with individual user preferences.

\end{abstract}

\section{Introduction}

Text-to-image diffusion models have made remarkable progress in visual fidelity, prompt adherence, and compositional generation, powering increasingly capable systems such as Stable Diffusion~\citep{podell2024sdxl, rombach2022high}, DALL-E~\citep{ramesh2021zero}, and Seedream \citep{seedream2025seedream}. To adapt these general-purpose models to downstream requirements, prior work has explored supervised or reinforcement-learning-based fine-tuning, the injection of additional conditioning signals, and reward-guided inference \citep{wallace2024diffusion, prabhudesai2023aligning}. These approaches have proven effective for applications including subject-driven generation, style customization, and general preference alignment.

Despite their effectiveness in many applications, personalizing diffusion models to individual users remains challenging, as it requires accurately modeling each user's preferences from limited feedback.
Existing approaches generally obtain user preferences in two ways: using vision-language models to summarize a user's historical preference records into a profile or embedding~\citep{dang2025personalized,kim2025draw}, or collecting explicit human annotations of users' likes and dislikes and converting them into structured visual preferences to guide generation~\citep{salehi2024viper}.
These approaches have two important limitations. First, incorporating preference representations requires additional fine-tuning of the diffusion model or training of a specialized adapter, introducing substantial computational cost and limiting scalability across different generative models. Second, acquiring reliable preference representations remains difficult: profiles inferred by vision-language models may be inaccurate when user histories are short, long, or heterogeneous, while collecting detailed human annotations imposes considerable burden on users.

In this work, we propose \ourmech (\textbf{P}ersonalized \textbf{a}t \textbf{T}est Time for diffusion models), which instantiates a personalized reward model at test time directly from a user's historical preference feedback. For each pair of images generated for a prompt, a user only needs to indicate which image they prefer. 
Our approach assumes that user preferences admit a low-dimensional latent representation, with individual differences captured by the weights each user assigns to shared preference dimensions.
We therefore formulate personalization as a two-stage problem: first, identifying a compact, shared visual attribute space that captures diverse preference-relevant attributes; and second, quantifying each user's preferences along these dimensions.
To identify the shared preference dimensions, we draw on the collection of over 500 visual attributes compiled by~\citet{salehi2024viper}. While these attributes provide broad coverage of visual preferences, estimating a separate weight for every attribute would require fitting hundreds of user-specific parameters from only a small number of pairwise observations, making preference estimation prone to overfitting. We therefore seek a compact shared representation that preserves meaningful variation across visual attributes while reducing the number of parameters to be estimated for each user. To this end, we use an autoencoder to derive a low-dimensional set of shared preference dimensions grounded in these attributes. Unlike covariance-based linear methods such as PCA ~\citep{pearson1901liii,hotelling1933analysis} and factor analysis\citep{spearman1961general,thurstone1931multiple}, the autoencoder can capture a nonlinear low-dimensional representation that better preserves interactions among visual attributes. The advantage is supported by our empirical results in Table~\ref{tab:dimension-reduction}. 
An evaluator then maps a prompt--image pair $(c,x)$ to its scores $s_{\theta}(c,x)\in\mathbb{R}^{d}$ along these dimensions, providing a shared basis for evaluating the visual characteristics of images.

To capture each user's individual preferences, we construct a personalized reward model that can be efficiently learned at test time. We quantify the reward of user $u$ as a linear combination $r_u(c,x)=\lambda_u^{\top}s_{\theta}(c,x)$, where $\lambda_u$ is the user-specific preference weight.
Given the user's preference history represented as a collection of $n$ prompt--image preference pairs ${(c_k,x_k^{+},x_k^{-})}_1^n$, we estimate the user-specific weights using the Bradley--Terry objective,
\begin{equation}
\lambda_u^{*}
=\arg\max_{\lambda_u}
\sum_{k=1}^{m_u}
\log \sigma\left(
\lambda_u^{\top}
\left[s_{\theta}(c_k,x_k^{+})-s_{\theta}(c_k,x_k^{-})\right]
\right).
\end{equation}
This factorization reduces test-time personalization to estimating a low-dimensional weight vector, keeping per-user optimization compact and enabling data-efficient learning from sparse feedback. The resulting reward is used to guide generation at inference time while the underlying diffusion model remains frozen. To improve generalization for users with very few preference pairs, we further introduce a mixed-effects parameterization $\lambda_u=G+b_u$, where $G$ captures population-level preferences shared across users and $b_u$ captures user-specific deviations. By pooling preference information through $G$, this formulation further enables more reliable personalization in few-shot settings. We evaluate our approach on real-user preference data, achieving approximately $78\%$ held-out pairwise preference prediction accuracy. When transferred to personalized image generation, our method achieves a $63.0\%$ win rate against the non-personalized generator. Our main contributions are as follows:
\begin{itemize}
\item We introduce the first mathematical formulation of user profiling for personalized image generation based directly on offline pairwise preferences. Our formulation represents each user by a low-dimensional preference vector over shared visual dimensions, turning personalization into a tractable parameter-estimation problem. 
\item We propose \ourmech, an end-to-end pipeline that fits a compact personalized reward and uses it to guide a frozen diffusion model through test-time noise optimization, requiring neither per-user model fine-tuning nor separate personalized generators.
\item We empirically demonstrate that our approach accurately models real users' preferences under limited feedback and that the learned personalized rewards transfer effectively from preference prediction to personalized image generation.
\end{itemize}

\begin{figure}
    \centering
    \includegraphics[width=\linewidth]{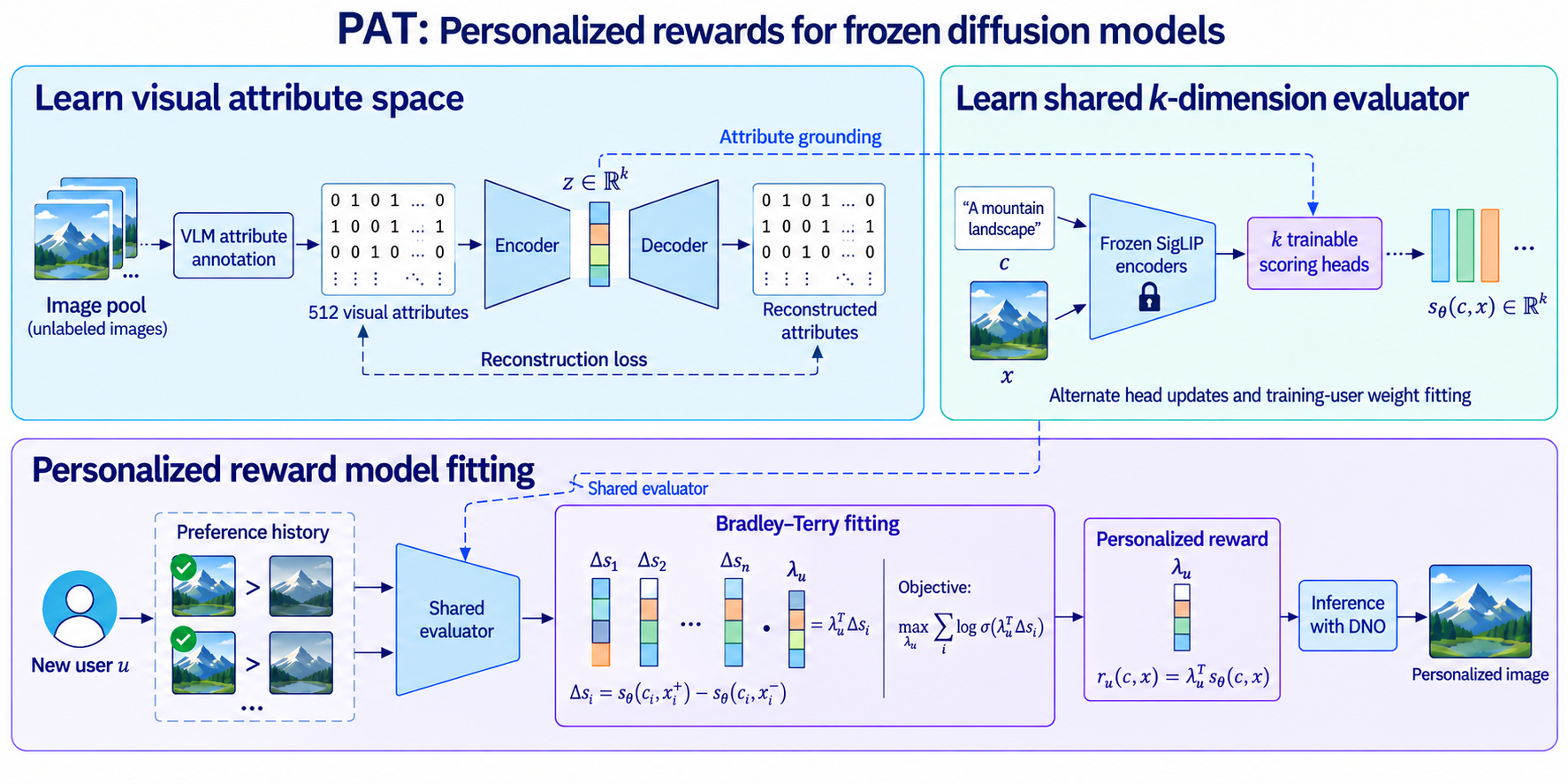}
    \caption{\textbf{Overview of PAT.} We learn a shared low-dimensional visual preference space and evaluator, fit a personalized reward from each user's pairwise preference history, and use the reward to guide a frozen diffusion model at inference time.}
    \label{fig:pat}
\end{figure}

\section{Related Work}
\paragraph{Diffusion Model Alignment.}

A growing body of work aligns pretrained diffusion models with downstream objectives such as human preferences, aesthetic quality, and safety-awareness. One class of approaches fine-tunes the diffusion model using reward supervision, thereby incorporating the desired objective into the model parameters~\citep{dai2023emu,  betker2023improving, clark2023directly, prabhudesai2023aligning, fan2023dpok, black2023training}. Another class performs alignment at inference time, steering the denoising trajectory or directly optimizing the latent variables while keeping the pretrained diffusion model frozen~\citep{jiang2026rethinking, goyal2026unified,tang2024tuning, yeh2024training, kimtest, zhai2025mira}. More recently, reinforcement learning has been introduced to diffusion model alignment by treating denoising as a sequential decision-making process. Existing methods employ policy-optimization algorithms such as GRPO~\citep{zheng2026diffusionnft, liu2026flow, xue2025dancegrpo,li2025mixgrpo, li2025uniworld, li2026aegpo}, as well as RLHF and direct preference optimization objectives such as DPO~\citep{wallace2024diffusion, jiang2026rethinking, liang2024step,wang2025diffusion, lee2025calibrated, gu2024diffusion, D3PO, dspo}, to improve generation according to human feedback or learned rewards. While most of these methods perform alignment by updating the diffusion model during training, a small number of concurrent works, besides ours, have also explored alignment at inference time. ~\citet{jin2025inference} and ~\citet{xie2026test} investigate training-free inference-time reward transfer from RL aligned-base difference between tuned model and base model. \citet{chen2026test}  treats intermediate flow latents as implicit policies and optimizes them at test time using PRDP and GRPO objectives.

\paragraph{Personalization for Generative Models.}

Personalization has been extensively studied for large language models through approaches such as user-profile conditioning, memory augmentation, parameter-efficient adaptation, and personalized reward modeling~\citep{yang2024rewards,li2026aplaud,poddar2024personalizing,wang2024conditional,song2026card,yu2026exact,wang2026learning,wu2026prefreward,tan2026instant,barreto2025capturing,cai2026one,iyerpersonalized,chen2025pad,shenfeld2025language,kim2025drift}. 
Personalizing diffusion models poses additional challenges. An LLM can jointly interpret heterogeneous user histories, extract preference-relevant information, and incorporate the resulting profile into generation through natural-language conditioning. In contrast, a diffusion model primarily serves as a generator and does not natively provide a comparable interface for reasoning over user histories or constructing explicit user models. Existing approaches therefore commonly rely on external vision-language models to summarize a user's prior choices or encode them into preference representations, while others use human-defined attributes or manual annotations to characterize the relevant dimensions of variation~\citep{dang2025personalized,lee2025personalized,salehi2024viper,kim2025draw,wang2026premier,li2026instant}. Our approach makes diffusion personalization more scalable by directly learning shared preference dimensions and user-specific rewards from pairwise feedback. It requires neither external VLM-based nor manual analysis of each user's history, and personalizes generation at inference time without fine-tuning the underlying diffusion model.

\section{Methodology}
\subsection{Problem Formulation}

For each user $u$, we observe a small set of offline preference pairs, $\mathcal{D}_u = \{(c_i,x_i^+,x_i^-)\}_{i=1}^{m_u}$,
where in $i$-th image pair, $c_i$ is a text prompt and $x_i^+$ is preferred to $x_i^-$. We assume that user preferences lie in a shared $k$-dimensional space, represented by an evaluator
$s(c,x)\in\mathbb{R}^k$. The personalized reward is a linear combination
\begin{equation}
    r_u(c,x) = \lambda_u^\top s(c,x),
    \qquad \lambda_u\in\mathbb{R}^k,
    \label{eq:personalized-reward}
\end{equation}
where $\lambda_u$ is the only user-specific parameter. At inference time, we aim to generate images maximizing the personal reward $r_u(c,x)$. 

\subsection{Learning an AutoEncoder Preference Evaluator}

\paragraph{Preference Dataset.}
We construct the data for learning the shared preference evaluator from the training split of Pick-a-Pic v1 \citep{kirstain2023pick}, where each comparison consists of a text prompt $c$, two generated images $(x^{+},x^{-})$, and the image preferred by the user. We sample a dimension-learning pool of 50,000 images from light raters. To obtain structured supervision describing the visual properties of each image, we annotate the pool with Qwen2.5-VL-72B \citep{bai2025qwen25vltechnicalreport} using the ViPer attribute taxonomy \citep{salehi2024viper}. The taxonomy contains approximately 512 attributes organized into 21 visual categories, including art style, color palette, composition, and mood. For each image, Qwen2.5-VL-72B assigns a binary label indicating whether an attribute is present in the image. Aggregating these annotations produces a binary attribute matrix with size $50{,}000\times512$, which is subsequently used to train the autoencoder.

\paragraph{Autoencoder for dimensionality reduction.}

Assigning a separate preference weight to each of the 512 attributes would require estimating a high-dimensional user vector from limited feedback, making reliable estimation difficult. We therefore learn a shared representation with $k\ll512$ dimensions, reducing the number of user-specific parameters to estimate. The bottleneck dimension $k$ controls the trade-off between representing diverse preferences and learning them from sparse feedback. To effectively reduce dimension, we use an autoencoder to capture potentially nonlinear attribute dependencies. Specifically, let $a_i\in\{0,1\}^{512}$ denote the attribute vector of image $x_i$ in the annotated pool of $N=50{,}000$ images. An encoder $E_{\eta}$ produces a latent representation, and a decoder $D_{\psi}$ reconstructs the attribute probabilities:
\begin{equation}
    z_i=E_{\eta}(a_i)\in\mathbb{R}^k,
    \qquad \hat a_i=D_{\psi}(z_i)\in(0,1)^{512}.
    \label{eq:ae-representation}
\end{equation}
We learn the encoder and decoder parameters $(\eta,\psi)$ by minimizing
\begin{equation}
    \mathcal{L}_{\mathrm{AE}}
    =\frac{1}{N}\sum_{i=1}^{N}
    \ell_{\mathrm{rec}}\!\left(a_i,\hat a_i\right),
    \label{eq:ae-reconstruction}
\end{equation}
where $\ell_{\mathrm{rec}}$ is the binary cross-entropy loss across attributes. After training, we discard the decoder and use $z_i$ as shared supervision targets for the  evaluator $s_{\theta}(c_i,x_i)$. These compact scores encode visual attribute structure rather than user preferences; each user's preferences are subsequently learned through the weights $\lambda_u$ over the evaluator's $k$ outputs.

\paragraph{Evaluator for each dimension.}
We train a differentiable evaluator $s_{\theta}(c,x)\in\mathbb{R}^{k}$ that combines general human preference learning with dimension-specific visual attribute supervision. A frozen SigLIP model encodes the caption and image, whose representations are concatenated and passed to MLP scoring heads with output dimension of $k$. Following the pairwise preference-learning principle of PickScore~\citep{kirstain2023pick}, each head learns to assign higher scores to human-preferred images. We additionally supervise each head using the corresponding autoencoder latent dimension: for each dimension, we supervise the top 33\% of pairs by absolute latent difference using the sign of that difference as the ordering target, then apply an attribute-ranking loss only to image pairs with different binary labels in that dimension. This auxiliary objective encourages the heads to retain distinct visual attribute signals while learning general preferences. We train the scoring heads on around 36k image pairs while keeping the SigLIP encoders and autoencoder fixed. The resulting scores form a shared basis for personalized reward fitting. Details are provided in Appendix~\ref{app:evaluator-train}.

\subsection{Personalized Reward Learning}

Given the shared evaluator and user $u$'s offline preference pairs
$\mathcal{D}_u$, we infer the weight vector $\lambda_u$ that best explains the user's
choices. Under the Bradley--Terry model, the probability of preferring
$x_i^+$ to $x_i^-$ is
\begin{equation}
    P_u(x_i^+\succ x_i^-\mid c_i)
    = \sigma\!\left(r_u(c_i,x_i^+)-r_u(c_i,x_i^-)\right)
    = \sigma\!\left(\lambda_u^\top\Delta s_i\right),
    \label{eq:bt-preference}
\end{equation}
where $\sigma(a)=(1+e^{-a})^{-1}$ and
$\Delta s_i=s(c_i,x_i^+)-s(c_i,x_i^-)$. The basic regularized
maximum-likelihood estimator is
\begin{equation}
    \hat\lambda_u
    = \arg\max_{\lambda\in\mathbb{R}^k}
    \left\{\sum_{i=1}^{m_u}\log\sigma(\lambda^\top\Delta s_i)
    -\gamma\|\lambda\|_2^2\right\},
    \label{eq:bt-basic}
\end{equation}
where $\gamma$ controls regularization and $m_u$ is the number of total preference pairs for user $u$. This formulation reduces personalization to estimating a low-dimensional vector, but its effectiveness still depends on the shared evaluator providing meaningful preference
coordinates.

\paragraph{Alternating evaluator and weight optimization.}
The collected image preference pairs can further train the shared evaluator
to better reflect preferences across users. We alternate between fitting
each user's $\lambda_u$ with the evaluator fixed and updating the evaluator
heads with the user weights fixed. Repeating these two steps allows the
shared preference representation and user-specific weights to adapt together,
while the evaluator's image and text encoders and the diffusion model remain
frozen.

\paragraph{Mixed-effects preference modeling.}
We further extend Equation~\ref{eq:bt-basic} to a mixed-effect  model, parameterizing each user's weights as
$\lambda_u=G+b_u$. Here, $G\in\mathbb{R}^k$ is a shared fixed effect
capturing population-level preferences, and $b_u\in\mathbb{R}^k$ is a
user-specific random effect capturing deviations from this shared component.
With an isotropic Gaussian prior $b_u\sim\mathcal{N}(0, I)$, 
we jointly estimate the shared component and user deviations by maximizing
the regularized log-likelihood:
\begin{equation}
    \begin{aligned}
    (\hat G,\{\hat b_u\})
    = \arg\max_{G,\{b_u\}}\; 
    \sum_{u\in\mathcal{U}_{\mathrm{train}}}\sum_{i=1}^{m_u}
    \log\sigma\!\left((G+b_u)^\top\Delta s_{u,i}\right)
    -\gamma_{b}
    \sum_{u\in\mathcal{U}_{\mathrm{train}}}\|b_u\|_2^2,
    \end{aligned}
    \label{eq:mixed-effects-bt}
\end{equation}
where $\mathcal{U}_{\mathrm{train}}$ denotes the set of
 training users,
and $\gamma_{b}$ controls the allowed
user-level variation, and $m_u$ is the number of total preference pairs for user $u$. This mixed-effect model is particularly useful when each user provides only
a few preference pairs. The shared component
$G$ is learned across users, allowing data-sparse users to benefit from
population-level evidence, while the penalty on $b_u$ limits unsupported
individual deviations. This can reduce the variance of user-specific
estimates and improve preference prediction under limited feedback.

\subsection{Inference-Time Generation with Direct Noise Optimization}

After learning the personalized reward $r_u$, we use Direct Noise Optimization (DNO) \citep{tang2024tuning} to guide generation without updating the diffusion model. Let $M_{\phi}(c,x_T)$ denote the image produced by the frozen diffusion model $\phi$ from prompt $c$ and initial noise $x_T\sim\mathcal{N}(0,I)$. 
Given a user's personal reward model $r_u(c,x)$, we aim to find the locally optimal noise $x_T^\star$ such that 
\begin{equation*}
    x_T^\star = \arg \max_{x_T} r_u(c, M_{\phi}(c,x_T)).
\end{equation*}
The optimal noise is derived from gradient ascent. At the $k$-th optimization step, the $k+1$ noise $x_T^{k+1}$ is updated by,
\begin{equation}
    x_T^{k+1} := x_T^k+ \ell \nabla_{x_T} r_u(c, M_{\phi}(c,x_T^k)).
\end{equation}
Note that while ~\citet{tang2024tuning} mainly uses an SDE sampler and optimizes over both initial noise and step-wise injected noises, we use ODE sampler thus only the initial noise is optimized. Additional implementation details are provided in Appendix~\ref{app:dno}.

\section{Experiments}
Our experiments evaluate two central questions: (1) whether the learned
personalized reward accurately represents a user's preferences, and (2) whether this reward can be used to generate images that match those preferences. We first study a controlled toy setting with manually specified synthetic preference profiles, and then extend the evaluation to real user preferences from Pick-a-Pic.

\subsection{Toy Example Setting}~\label{sec:toy-exp-sec}

\paragraph{Goal, data, and evaluation.}
We begin with a controlled toy experiment involving five synthetic users, each represented by a manually specified ground-truth preference vector $\lambda^\star$ (Table~\ref{tab:toy-configurations}). Their preferences are defined over five candidate reward dimensions:
\texttt{clip}~\citep{radford2021learning}, \texttt{aes}~\citep{AES},
\texttt{color}~\citep{hasler2003measuring}, \texttt{hps}~\citep{wu2023human}, and
\texttt{pickscore}~\citep{kirstain2023pick}. These scores capture complementary aspects of visual appearance and text--image alignment; details are provided in
Appendix~\ref{app:metrics-data}. We repeat the experiment three times, sampling 200
captioned image pairs from Pick-a-Pic v1 in each repetition and splitting them into 100 training and 100 test pairs. After standardizing each reward dimension over the sampled images, we construct synthetic preference labels by selecting the image with the higher ground-truth weighted reward, $(\lambda^\star)^\top r(c,x)$. We then estimate $\lambda$ from the training preferences by maximizing the Bradley--Terry likelihood with regularization 0.1. Evaluation considers the preference prediction accuracy on the held-out pairs. We compare the learned multi-dimensional reward with the ground-truth reward and a best one-hot baseline, which uses the highest-performing individual reward
dimension available in each configuration. This comparison tests whether
learning a combination of reward signals better captures a user's preferences than relying on a single evaluator.

\begin{table}[h]
    \centering
    \begin{minipage}[t]{0.38\textwidth}
        \vspace{0pt}
        
\begingroup
    \centering
    \captionof{table}{Held-out pairwise preference prediction accuracy(\%) for the synthetic
    configurations. Best one-hot uses the strongest individual reward dimension.
    The learned combination achieves higher accuracy.}
    \label{tab:toy-one-hot}
    \small
    \setlength{\tabcolsep}{4pt}
    \begin{tabular}{l c c}
        \toprule
        Config & \ourmech & Best one-hot \\
        \midrule
        Uniform & \textbf{90.0} $\pm$ 4.3 & 77.0 $\pm$ 1.6 \\
        Normal-1 & \textbf{90.7} $\pm$ 1.2 & 80.3 $\pm$ 3.3 \\
        Normal-2 & \textbf{88.7} $\pm$ 3.6 & 79.0 $\pm$ 3.7 \\
        Heavy-1 & \textbf{91.0} $\pm$ 5.7 & 81.7 $\pm$ 4.8 \\
        Heavy-2 & \textbf{90.7} $\pm$ 4.4 & 81.0 $\pm$ 5.4 \\
        \bottomrule
    \end{tabular}
\endgroup
    \end{minipage}\hfill
    \begin{minipage}[t]{0.60\textwidth}
        \vspace{0pt}
        \begingroup
    \centering
    \captionof{table}{Held-out pairwise preference accuracy (\%) on
    631 Pick-a-Pic users, grouped by the number of training pairs.
    Each user has 50 test pairs. Bold indicates the best accuracy
    in each group. }
    \label{tab:real-user-few-shot}

    \small
    \setlength{\tabcolsep}{1pt}

    \begin{tabularx}{\linewidth}{
        @{}
        >{\raggedright\arraybackslash}p{0.42\linewidth}
        *{4}{>{\centering\arraybackslash}X}
        @{}
    }
        \toprule
        Training pairs & 11--20 & 21--30 & 31--60 & 61$+$ \\
        \midrule
        \ourmech
        & \textbf{73.1} \tiny{$\pm$ 1.3} & \textbf{74.6} \tiny{$\pm$ 0.9} & \textbf{77.1} \tiny{$\pm$ 0.3} & \textbf{78.2} \tiny{$\pm$ 0.1} \\

        w/o Mixed Effect
        & \textbf{73.1} \tiny{$\pm$ 1.4} & 73.5 \tiny{$\pm$ 0.9} & 76.4 \tiny{$\pm$ 0.4} & 77.9 \tiny{$\pm$ 0.1} \\

        w/o Alternating Training
        & 71.2\tiny{$\pm$ 1.1} & 72.3\tiny{$\pm$ 0.9} & 75.2\tiny{$\pm$ 0.7} & 75.4\tiny{$\pm$ 0.2} \\

        w/o Alternating Training \&  Mixed Effect
        & 71.0\tiny{$\pm$ 1.0} & 71.6\tiny{$\pm$ 0.7} & 74.7\tiny{$\pm$ 0.7} & 74.7\tiny{$\pm$ 0.3} \\

        \midrule

        Universal $\lambda$
        & 67.6\tiny{$\pm$ 0.4} & 66.9\tiny{$\pm$ 0.6} & 69.3\tiny{$\pm$ 0.2} & 70.3\tiny{$\pm$ 0.3} \\

        w/ Frozen Evaluator
        & 66.9\tiny{$\pm$ 0.4} & 65.3\tiny{$\pm$ 0.6} & 65.1\tiny{$\pm$ 0.8} & 66.0\tiny{$\pm$ 0.4} \\
        \bottomrule
    \end{tabularx}
\endgroup
    \end{minipage}
\end{table}

\begin{figure}[h]
        \centering
        \includegraphics[width=\linewidth]{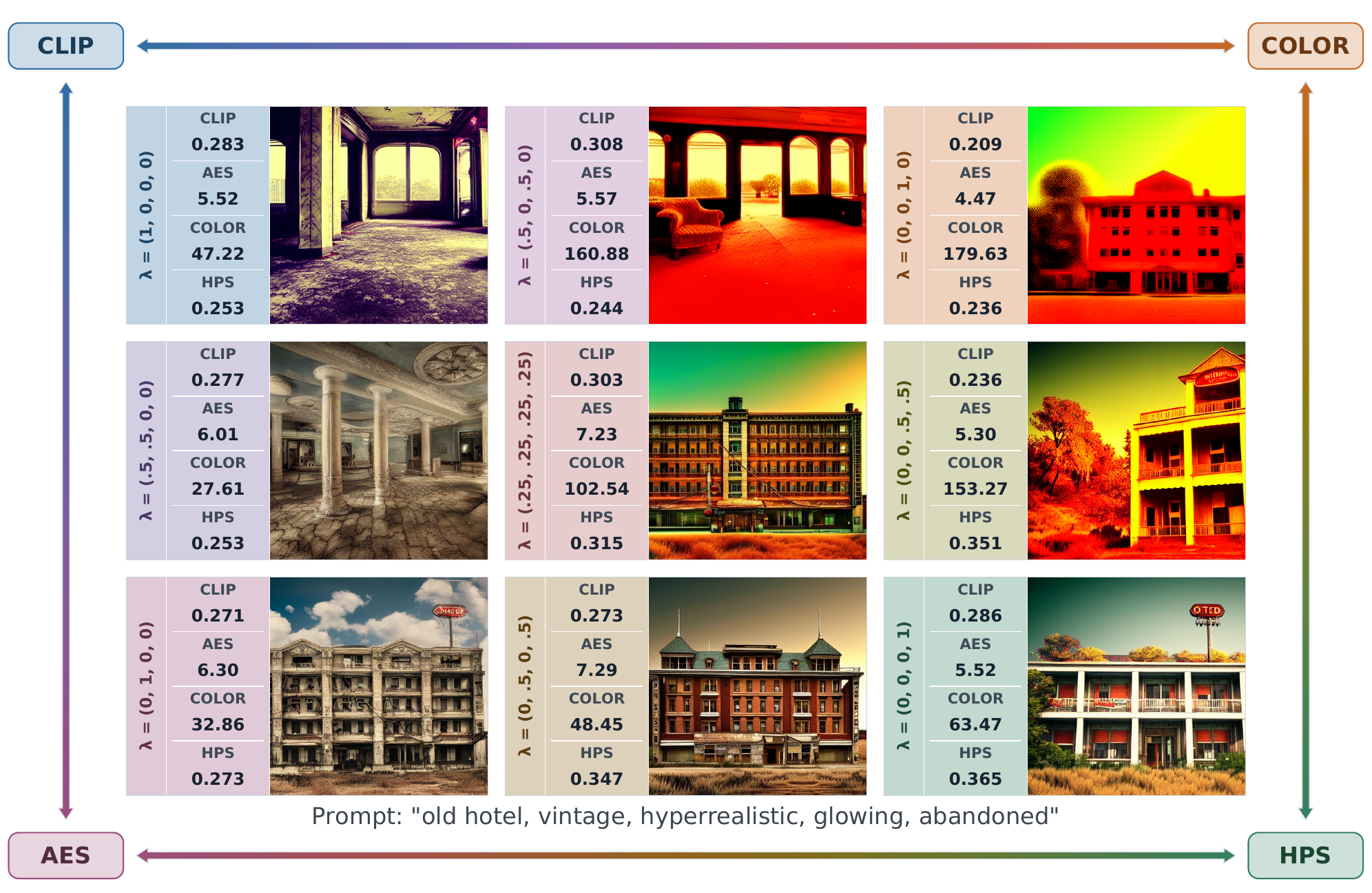}
        \caption{Qualitative examples of preference-controlled generation under a fixed prompt. Each image is generated using a different weight vector $\lambda$ over (\texttt{clip}, \texttt{aes}, \texttt{color}, \texttt{hps}), producing distinct visual styles. Corners emphasize individual reward dimensions, while intermediate positions combine multiple dimensions. The corresponding weights and metric scores are shown beside each image.}
        \label{fig:toy-example-show}
    \end{figure}

\paragraph{Results.} As shown in Table~\ref{tab:toy-one-hot}, our learned reward combination outperforms the best single reward dimension across all five synthetic preference configurations, achieving around 90\% held-out accuracy. These results demonstrate that fitting user-specific weights from limited pairwise feedback better captures diverse preference profiles than relying on any individual metric. Figure~\ref{fig:toy-example-show} qualitatively illustrates how different preference weights affect generation under the same caption. Increasing the weight on colorfulness produces more vivid, saturated images, while other weight configurations yield more muted palettes and different architectural compositions. These examples show that varying the preference weights enables different visual styles without changing the text prompt.

\subsection{Real User: Personalized Weight Modeling}

We next extend our evaluation from synthetic preferences to real user choices using the training split of Pick-a-Pic~\citep{kirstain2023pick}. We evaluate 631 users, holding out 50 preference pairs per user for testing and using their remaining pairs for training. To examine performance under different amounts of personal feedback, we groups users by their training-set sizes, with 66, 55, 120, and 390 users in the four groups, respectively. We measure held-out pairwise preference accuracy to assess how well each method predicts a user's unseen choices. We repeat the train-test split and model user preference for 3 times and report the average held-out pair test accuracy. We evaluate variants of \ourmech{} to examine the contribution of each pipeline component. The full method combines alternating evaluator and user-weight optimization, hierarchical preference modeling, and user-specific personalization. \emph{w/o mixed effect} replaces the joint parameterization $\lambda_u=G+b_u$ with separate per-user weight fitting regularized toward a fitted global preference vector, pushing back to an objective as Equation~\ref{eq:bt-basic}. \emph{w/o Alternating Training} freezes the shared evaluator during user-weight fitting. We also remove both components to evaluate their combined contribution. \emph{Universal $\lambda$} replaces user-specific weights with a shared population-level vector. Finally, \emph{Universal $\lambda$ w/ Frozen Evaluator} uses a fixed evaluator and a shared weight vector, removing all three components.

Table~\ref{tab:real-user-few-shot} shows that \ourmech{} achieves strong preference prediction performance across different amounts of user feedback. Personalization provides substantial gains: compared with using shared population-level weights, user-specific modeling improves accuracy by $5.5\%$--$7.9\%$ across all four groups, highlighting the importance of capturing individual preferences. Mixed-effects modeling is particularly useful when feedback is limited: it improves accuracy by $1.1\%$ and $0.7\%$ percentage points in the 20--30 and 30--60 groups, respectively, while matching independent fitting in the smallest group. Alternating training further improves accuracy across all groups, supporting joint adaptation of the evaluator and user-specific weights.

\begin{table}[t]
    \centering
    \begin{minipage}[t]{0.60\textwidth}
        \vspace{0pt}

\begingroup
    \centering
    \captionof{table}{Effect of autoencoder dimension $k$ on held-out pairwise
    preference accuracy (\%) for 631 users. Variants follow the component ablations
    in Table~\ref{tab:real-user-few-shot}.
    Bold indicates the best accuracy at each dimensionality.}
    \label{tab:real-user-main}

    \small
    \setlength{\tabcolsep}{4pt}
    \renewcommand{\arraystretch}{1.12}

    \begin{tabularx}{\linewidth}{
        @{}
        >{\raggedright\arraybackslash}p{0.38\linewidth}
        *{5}{>{\centering\arraybackslash}X}
        @{}
    }
        \toprule
        Variant & $k$=20 & $k$=40 & $k$=50 & $k$=80 & $k$=100 \\
        \midrule
        \ourmech
        & 75.28 & 76.88 & \textbf{77.14} & 77.17 & 77.10\\

        w/o Mixed Effect
        & 75.14  & 76.93 & 76.73 & 76.85 & 76.74\\

        w/o Alternating Training
        & 72.99  & 74.80 & 74.65 & 74.60 & 74.51\\

        w/o Alternating Training \& Mixed Effect
        & 72.60  & 74.24 & 74.04 & 74.23 & 74.36\\
        \midrule 
        Universal $\lambda$
        & 69.58  & 70.36 & 69.53 & 69.45 & 69.53\\

        w/ Frozen Evaluator
        & 64.72  & 65.88 & 65.86 & 65.51 & 65.47\\
        \bottomrule
    \end{tabularx}
\endgroup
    \end{minipage}\hfill
    \begin{minipage}[t]{0.37\textwidth}
        \vspace{0pt}
        \begingroup
    \centering
    \captionof{table}{General-purpose image-quality and prompt-alignment metrics for \ourmech{} and vanilla SD3.5. Win rate is the fraction of matched pairs in which our image receives a higher score.}
    \label{tab:image-quality-control}
    \setlength{\tabcolsep}{3pt}
    \renewcommand{\arraystretch}{1.12}
    \begin{tabular}{@{}l c c c@{}}
        \toprule
        Metric & Baseline & \ourmech & \shortstack{\ourmech \\win rate} \\
        \midrule
        PickScore & 0.2207 & 0.2203 & 47.6\% \\
        HPS & 0.2817 & 0.2817 & 52.0\% \\
        CLIP & 0.2825 & 0.2813 & 49.1\% \\
        AES & 5.913 & 5.856 & 45.4\% \\
        \bottomrule
    \end{tabular}
\endgroup
    \end{minipage}
\end{table}

\subsection{Ablation Studies}

\paragraph{Autoencoder vs.\ other dimension reduction methods.}
We compare the autoencoder with two linear dimension reduction methods:(a) a correlation-based baseline that groups correlated visual attributes and averages their binary labels to construct shared preference dimensions and (b) Principle Component Analysis(PCA) that uses SVD to factorize the binary label matrix and use the top $k$ principle dimension as shared preference dimensions. As detailed in Appendix~\ref{app:corr-ablation}, the autoencoder achieves highest held-out pairwise accuracy at every tested dimensionality. At $k=50$, accuracy improves from $75.39\%$ to $77.14\%$. These results support Autoencoder approach constructs a more effective basis for personalization than other linear dimension reduction methods.

\paragraph{Preference-space dimensionality.}
Table~\ref{tab:real-user-main} examines the effect of the autoencoder dimension $k$. Increasing $k$ from 20 to 50 improves \ourmech{} from $75.28\%$ to $77.14\%$ accuracy, suggesting that additional dimensions help capture heterogeneous user preferences. Beyond $k=50$, accuracy plateaus, reaching $77.17\%$ and $77.10\%$ at $k=80$ and $k=100$, respectively. These results suggest that 50 dimensions are sufficient to capture the preference variation needed for accurate prediction in this setting, with further increases providing no additional benefit for the full method.

\paragraph{Regularization strength.}
We vary the regularization coefficient $\gamma$ to examine its effect on preference estimation, with detailed results provided in Appendix~\ref{app:lambda-reg}. For the sake of notation brevity, we use $\gamma$ to represent the regularization strength including mixed-in effect.   Held-out pairwise accuracy peaks with $\gamma=0.03$. This trend suggests that moderate regularization helps generalization, whereas excessive regularization restricts the model's ability to capture user-specific preferences.

\begin{figure*}[h]
\centering
\setlength{\tabcolsep}{1.5pt} \begin{tabular}{ p{0.25\textwidth}  p{0.18\textwidth}  p{0.18\textwidth} p{0.18\textwidth} p{0.18\textwidth}}\toprule
\multicolumn{1}{c}{} & \multicolumn{1}{c}{} & \multicolumn{1}{c}{} & \multicolumn{1}{c}{} & \multicolumn{1}{c}{} \\

\multicolumn{1}{c}{\multirow{-2}{*}{\textbf{User Profile}}} &  
\multicolumn{1}{c}{\multirow{-2}{*}{\textbf{SD 3.5}}} & \multicolumn{1}{c}{\multirow{-2}{*}{\textbf{SD 3.5+Profile}}} & \multicolumn{1}{c}{\multirow{-2}{*}{\textbf{PPD}}} &  \multicolumn{1}{c}{\multirow{-2}{*}{\textbf{PAT}}} \\

\midrule
 
\multirow[t]{2}{0.25\textwidth}{\raggedright\footnotesize
Visually, they favor clean, centered compositions with strong subject isolation, \textcolor{Green}{soft natural textures, warm light, lush greens, creamy whites, and controlled saturated accents}.} &
\includegraphics[scale=0.135,valign=t]{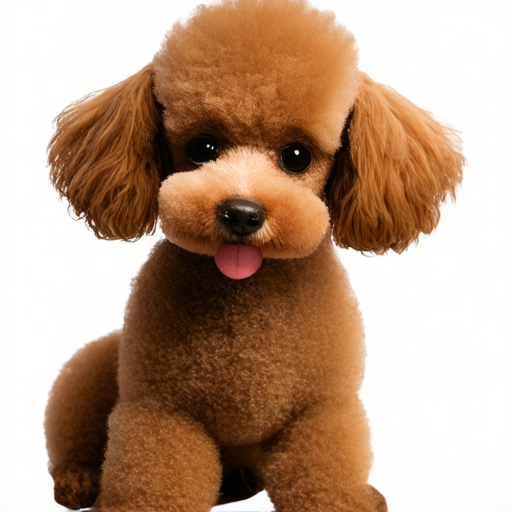} \centering
&  \includegraphics[scale=0.135,valign=t]{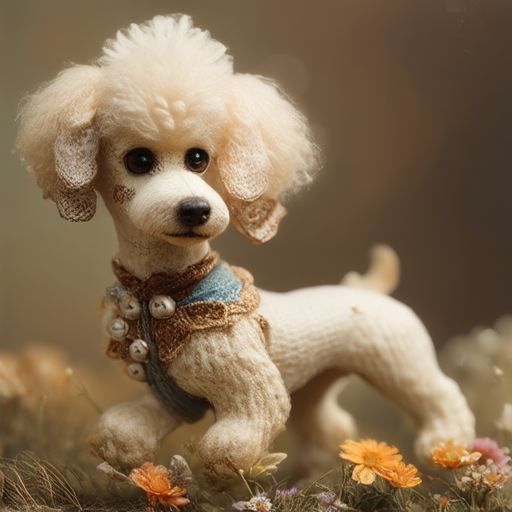} \centering
& \includegraphics[scale=0.135,valign=t]{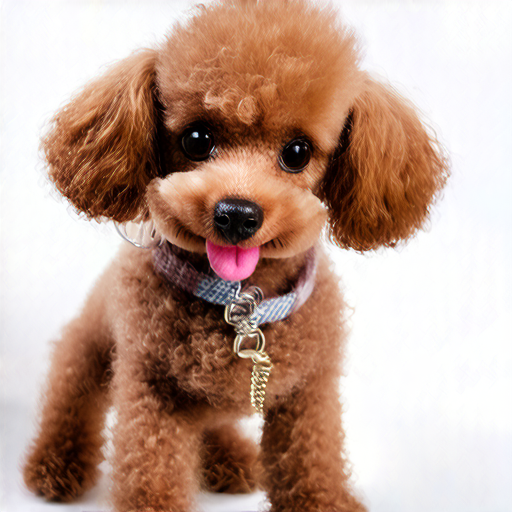} \centering
& \includegraphics[scale=0.135,valign=t]{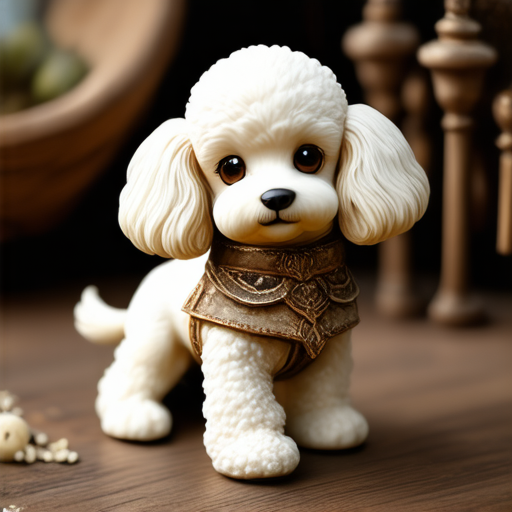} \\
& \multicolumn{4}{c}{\emph{\small a mature poodle toy}}  \\ \midrule

\multirow[t]{2}{0.25\textwidth}{\footnotesize 
Visually, they prefer \textcolor{Green}{earthy textures, architectural intricacy, and compositions} ... consistently reject \textcolor{red}{overly glossy, cartoonish, exaggerated, or poster-like imagery; extreme close-ups, simplified fantasy shapes}...}
&  \includegraphics[scale=0.135,valign=t]{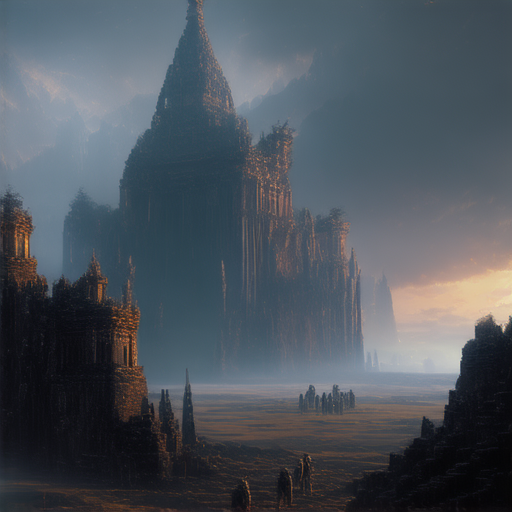} 
&  \includegraphics[scale=0.135,valign=t]{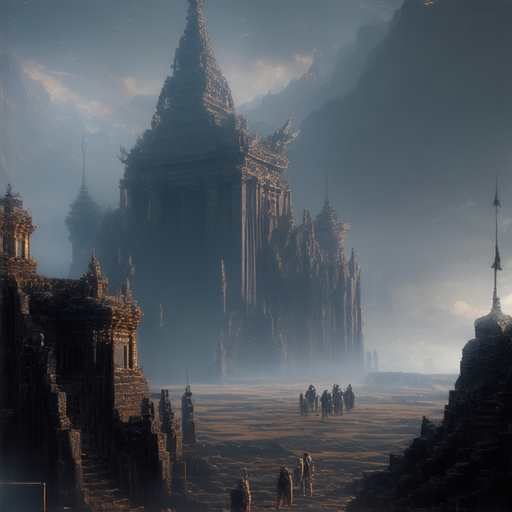}  
& \includegraphics[scale=0.135,valign=t]{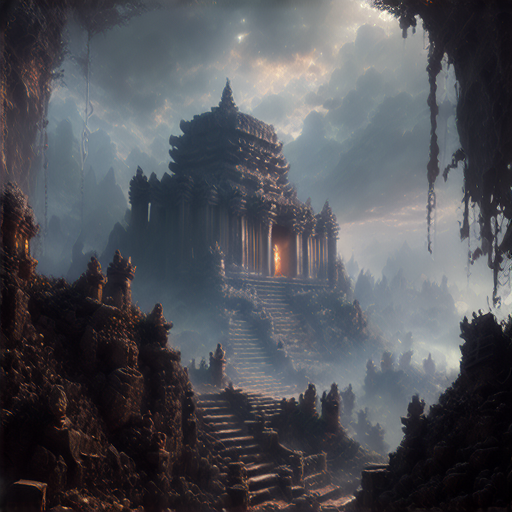} 
& \includegraphics[scale=0.135,valign=t]{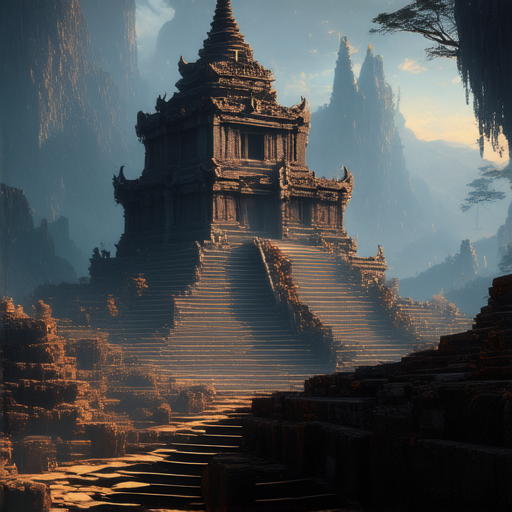} \\
& \multicolumn{4}{c}{\emph{\small a Majestic Ancient Temple, Moody, Fantasy Landscape, Volumetric Lighting}}  \\ 
& \multicolumn{4}{c}{\emph{\small Digital painting, Marc Allante, Studio Ghibli, dramatic, Intricate detail}}  \\ \midrule

\multirow[t]{2}{0.25\textwidth}{\footnotesize 
They like refined palettes, stylized realism, clean graphic shapes, purposeful costume/armor design, and \textcolor{Green}{compositions that place the subject in a meaningful environment. }}
&  \includegraphics[scale=0.135,valign=t]{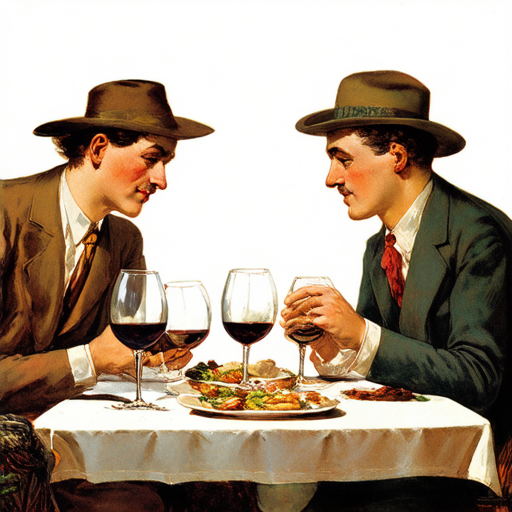} 
&  \includegraphics[scale=0.135,valign=t]{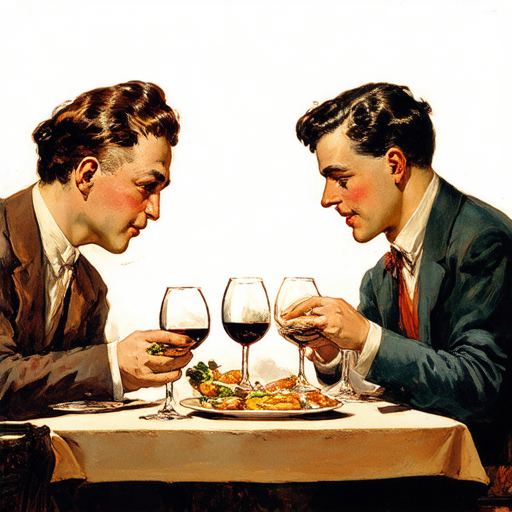}  
& \includegraphics[scale=0.135,valign=t]{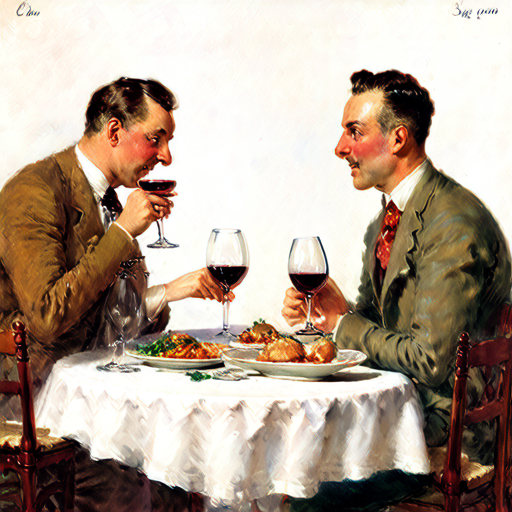} 
& \includegraphics[scale=0.135,valign=t]{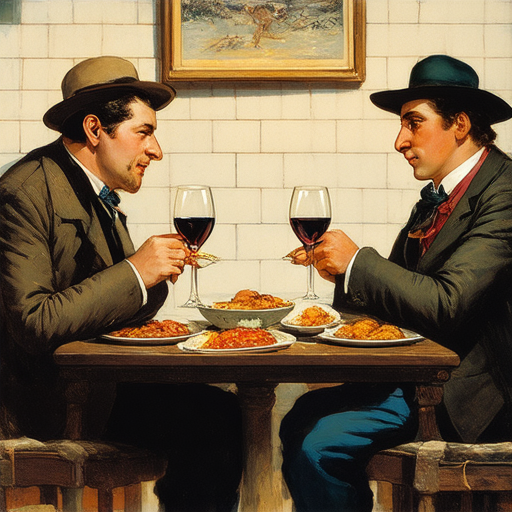} \\
& \multicolumn{4}{c}{\emph{\small Illustration of two men sitting at a table eating food and drinking wine}}  \\ \midrule
\multirow[t]{2}{0.25\textwidth}{\footnotesize 
Aesthetically, they favor \textcolor{Green}{clean, centered compositions with a strong iconic subject}, ... reject \textcolor{red}{cluttered scenes, busy narrative cartoons, muddy or overly literal landscapes}.}
 &  \includegraphics[scale=0.135,valign=t]{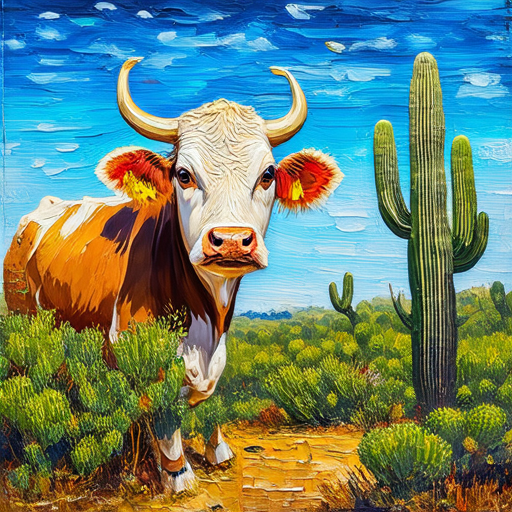} 
&  \includegraphics[scale=0.135,valign=t]{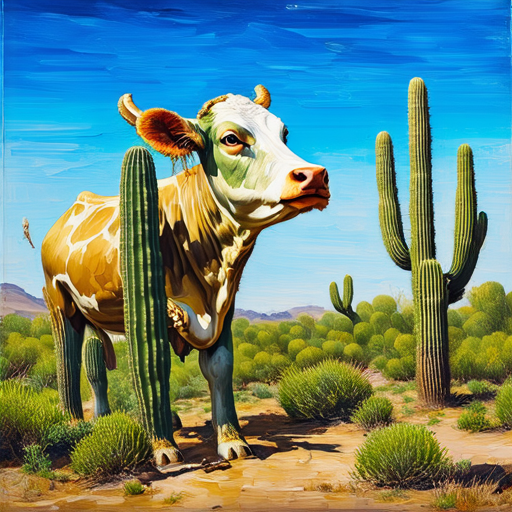}  
& \includegraphics[scale=0.135,valign=t]{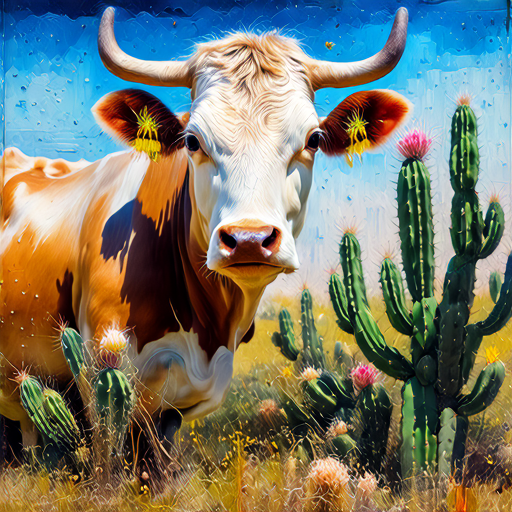} 
& \includegraphics[scale=0.135,valign=t]{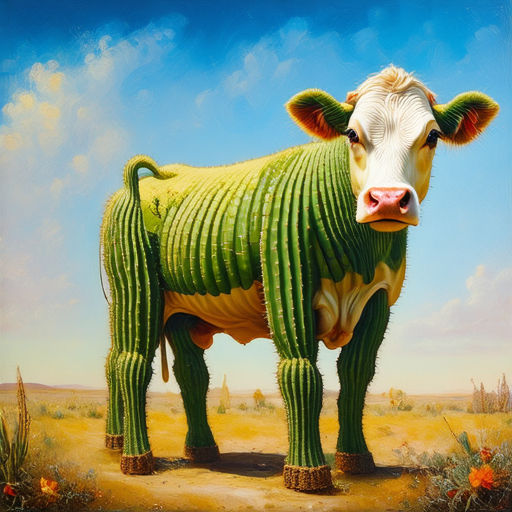} \\
& \multicolumn{4}{c}{\emph{\small Cow made of Cactus oil painting}}  \\ 
\bottomrule
    \end{tabular}
\caption{\textbf{Qualitative comparison of personalization methods.} For each user, we compare images generated by vanilla SD3.5, Profile-as-Prompt, PPD, and PAT under the same prompt. We highlight the user preferences in \textcolor{Green}{green} and dispreferences in \textcolor{red}{red}. More qualitative results are in Appendix~\ref{app:more-qualitative}.}
    \label{fig:user_profile_qualitative}
\end{figure*}

\subsection{Real User: Personalized Image Generation}

\paragraph{Experimental setup.}
We evaluate whether the learned personalized rewards improve image generation beyond ranking existing images. We include all 631 users and generate 10 images per user for each method. We use Stable Diffusion 3.5 (SD3.5) as the base generator, with 30 sampling steps and a classifier-free guidance scale of 4.5. For \ourmech{}, we optimize the initial noise using Direct Noise Optimization (DNO) with the learned personalized reward, while keeping the diffusion model frozen. DNO runs for 30 optimization steps with a learning rate of 0.005. We use GPT-5.5 as the judge under a blind, profile-aware evaluation protocol, comparing each method's outputs against vanilla SD3.5 for alignment with the user's preferences. latex
We first provide GPT-5.5 with 10 image pairs and the user's recorded preferences, one pair for each test prompt, to construct a user preference profile. We then use this profile to guide GPT-5.5 in judging which generated image better aligns with the user's preferences. The prompt templates are provided in Appendix~\ref{app:prompts}.

\paragraph{Baselines.}
We compare against our reproduction of Personalized Preference Fine-tuning of Diffusion Models (PPD). For each user, PPD randomly selects four historical prompt--image pairs and uses a 72B Qwen vision-language model to extract the user's preferences. The resulting VLM embeddings serve as additional conditioning signals for diffusion model fine-tuning, and the texts summarized from VLM are treated as user profile. At generation time, we set PPD's user-signal guidance scale to 1. We also include a \emph{Profile-as-Prompt} baseline, which appends the textual user profile produced by PPD's VLM to the original generation prompt, directly conditioning SD3.5 on the preference description.

\paragraph{Generation results.}
Images generated by \ourmech{} are preferred over vanilla SD3.5 in \textbf{63.0\%} of comparisons, compared with 58.5\% for PPD and 55.7\% for Profile-as-Prompt. Our method therefore improves the win rate against the shared baseline by 4.5\% and 7.3\%, respectively. These results demonstrate that personalized rewards learned from offline preference pairs transfer effectively to generation-time optimization, achieving the highest win rate among the evaluated methods while keeping the diffusion model frozen.

\paragraph{Image-quality control.}
To assess whether the personalized generation gains reflect a general improvement in image quality, we compare \ourmech{} with vanilla SD3.5 over the image pairs generated from the same prompts. We evaluate each pair using four general-purpose quality and alignment metrics: PickScore, HPS, CLIP, and AES. As shown in Table 4, average scores remain similar across methods, while metric-based win rates range from $45.4\%$ to $52.0\%$, showing no consistent advantage for \ourmech{} on these metrics. Together with the $63.0\%$ win rate under profile-aware evaluation, these results suggest that our gains reflect better alignment with individual preferences rather than a broad increase in generic image quality.

\section{Conclusion}

We introduced \ourmech{}, a framework for personalizing diffusion models at test time from historical pairwise preferences. By combining shared, attribute-grounded reward dimensions with user-specific weights, \ourmech{} enables preference estimation from limited feedback and guides generation while keeping the diffusion model frozen. Experiments demonstrate improved preference prediction and better personalized generation win rate over vanilla SD3.5, without consistent gains in generic quality metrics. These results support compact personalized rewards as an effective interface between individual preferences and inference-time image generation.

\subsection*{AI Use Statement}

We used generative AI tools to assist with manuscript editing, organizing experimental results, preparing figures, providing feedback on research methodology and experiments, and summarizing existing literature. We also used LLMs to assist with implementing our method and employed a vision-language model as a judge in part of our experimental evaluation. Generative AI tools were not used to originate research ideas, formulate the methodology, design experiments, or search for related literature. The authors reviewed all AI-assisted work and verified the correctness of LLM-generated code. We take responsibility for the final content of this work, including all text, claims, results, and artifacts produced with the assistance of generative AI.

\section*{Ethics Statement}
This work uses existing public preference data from Pick-a-Pic and involves no new participant recruitment or manual preference annotation. Automated visual attribute annotations and model-based generation evaluations may reflect biases in the underlying models and data. Personalized generation may reinforce biased preferences or produce harmful content, while user histories and learned preference vectors may reveal sensitive information. Practical deployment should therefore protect preference data and retain appropriate content-safety safeguards.

\section*{Reproducibility Statement}
Appendix~\ref{app:evaluator-train} details the evaluator architecture and training objectives, Appendix~\ref{app:evaluator-train} provides dataset descriptions, reward metrics, and ablation details. We will publicly release our code and experimental configurations to support reproduction and further research.

\bibliography{references}
\bibliographystyle{references}

\newpage
\appendix

\section{Evaluator Training}
\label{app:evaluator-train}

\paragraph{Architecture and supervision.}
Our evaluator produces a vector of scores
$s_{\theta}(c,x)=[s_{\theta,1}(c,x),\ldots,s_{\theta,k}(c,x)]^{\top}$
for a caption--image pair $(c,x)$. We encode the caption and image using frozen SigLIP encoders and concatenate their representations into a 2,304-dimensional feature vector(1152 dimension for both image and caption embeddings). This vector is passed through an MLP consisting of a linear layer with 512 hidden units, a GELU activation, and a linear output layer producing $k$ scalar scores. The output dimensions share the hidden representation but have separate output weights. We train the MLP on approximately 36,000 image pairs using two sources of supervision: human pairwise preferences and dimension-specific targets derived from the pretrained autoencoder. The former provides a common preference-learning objective across dimensions, while the latter encourages distinct visual attribute responses.

\paragraph{Human preference supervision.}
Let $(c_i,x_i^+,x_i^-)$ denote a training pair, where $x_i^+$ is preferred to $x_i^-$ by the user. For dimension $j$, we define the predicted score difference as
\begin{equation}
    d_{ij}
    =
    s_{\theta,j}(c_i,x_i^+)
    -
    s_{\theta,j}(c_i,x_i^-).
\end{equation}
Following the pairwise preference-learning principle of PickScore~\citep{kirstain2023pick}, we encourage each output dimension to assign a higher score to the preferred image. We use the binary cross-entropy loss
\begin{equation}
    \ell_{\mathrm{BCE}}(d,y)
    =
    -y\log\sigma(d)
    -(1-y)\log\bigl(1-\sigma(d)\bigr),
\end{equation}
where $d\in\mathbb{R}$ is a predicted score difference, $y\in\{0,1\}$ indicates whether the first image should receive a higher score than the second, and $\sigma$ is the sigmoid function. Since each pair is ordered by human preference, the target for this objective is always $1$. For a minibatch $\mathcal{B}$, the preference loss is
\begin{equation}
    L_{\mathrm{prefer}}
    =
    \frac{1}{|\mathcal{B}|k}
    \sum_{i\in\mathcal{B}}\sum_{j=1}^{k}
    \ell_{\mathrm{BCE}}(d_{ij},1).
\end{equation}

\paragraph{Visual attribute supervision.}
To construct dimension-specific supervision, we use the fixed encoder of the pretrained autoencoder:
\begin{equation}
    z_i^+=E_{\eta}(a(x_i^+)),
    \qquad
    z_i^-=E_{\eta}(a(x_i^-)),
\end{equation}
where $a(x)\in\{0,1\}^{512}$ is the binary visual attribute annotation of image $x$, $E_{\eta}$ is the autoencoder encoder with fixed parameters $\eta$, and $z_i^+,z_i^-\in\mathbb{R}^{k}$ are the corresponding continuous latent representations.

We first compute the latent difference for each pair and dimension:
\begin{equation}
    \delta_{ij}=z_{ij}^{+}-z_{ij}^{-}.
\end{equation}
For each dimension $j$, we rank the training pairs by $|\delta_{ij}|$ and retain the top 33\% for attribute supervision. Let $\mathcal{I}_j$ denote the indices of the retained pairs. We define the supervision mask and binary ordering target as
\begin{equation}
    m_{ij}=\mathbf{1}[i\in\mathcal{I}_j],
    \qquad
    y_{ij}^{\mathrm{attr}}=\mathbf{1}[\delta_{ij}>0].
\end{equation}
Thus, pair selection is based on the magnitude of the continuous latent difference, while the binary target records its direction. We do not binarize the individual images' latent values. This procedure focuses attribute supervision on pairs with relatively large differences within each dimension.

For a minibatch $\mathcal{B}$, the masked attribute loss is
\begin{equation}
    L_{\mathrm{attr}}
    =
    \frac{
        \sum_{i\in\mathcal{B}}\sum_{j=1}^{k}
        m_{ij}\,
        \ell_{\mathrm{BCE}}(d_{ij},y_{ij}^{\mathrm{attr}})
    }{
        \max\left(1,\sum_{i\in\mathcal{B}}\sum_{j=1}^{k}m_{ij}\right)
    }.
\end{equation}
Pairs outside $\mathcal{I}_j$ contribute no attribute loss for dimension $j$, but still contribute to the human preference loss. We optimize the sum of the two losses:
\begin{equation}
    L_{\mathrm{eval}}
    =
    L_{\mathrm{prefer}}+L_{\mathrm{attr}}.
\end{equation}
Together, these objectives encourage the evaluator to balance general preference prediction with dimension-specific visual attribute supervision. Only the MLP parameters are updated; the SigLIP encoders and autoencoder remain fixed during this training stage. The autoencoder outputs define supervision masks and targets, whereas both losses operate on the evaluator's predicted score differences. The evaluator therefore learns continuous, attribute-grounded preference scores rather than directly reconstructing the autoencoder latent values. These scores are subsequently combined using user-specific weights to construct personalized rewards.

\section{More Details about Experiments}\label{app:details}

\subsection{Reward Metrics}
\label{app:metrics-data}
Our controlled experiments define synthetic user preferences over five reward dimensions: \texttt{clip}, \texttt{aes}, \texttt{color}, \texttt{hps}, and \texttt{pick}. These scores measure text--image alignment, visual aesthetics, colorfulness, and learned human preferences. Although the dimensions may be correlated, they provide distinct signals that can be combined to construct different preference profiles.

\paragraph{CLIP similarity (\texttt{clip}).}
CLIP~\citep{radford2021learning} learns a shared image--text embedding space through contrastive training on image--text pairs. We use the similarity between the embeddings of a generated image and its corresponding prompt to measure semantic alignment. Higher scores indicate stronger correspondence between the image content and the text description. This dimension captures prompt relevance, although it does not directly measure visual quality or individual user preferences.

\paragraph{Aesthetic score (\texttt{aes}).}
The aesthetic predictor~\citep{AES} estimates an image's aesthetic appeal using a prediction head applied to its CLIP image embedding. It is trained to predict human aesthetic ratings and provides an image-only quality signal, without conditioning on the prompt. This dimension captures visual appeal that may not be reflected by text--image similarity alone.

\paragraph{Colorfulness (\texttt{color}).}
The colorfulness metric of \citet{hasler2003measuring} quantifies image colorfulness using statistics of red--green and yellow--blue opponent channels. It combines the variability and mean magnitude of these color signals to produce an image-only score. Unlike the learned reward models, this metric measures a specific visual property without preference supervision. Higher values indicate more colorful images, which need not be preferred by every user; its contribution to the synthetic reward is determined by the user's weight vector.

\paragraph{Human Preference Score (\texttt{hps}).}
Human Preference Score v2~\citep{wu2023human} is a CLIP-based scoring model fine-tuned on the Human Preference Dataset v2 (HPD v2). Given a prompt and an image, it predicts a score intended to reflect human preferences over generated images. Compared with the original CLIP similarity, this score incorporates explicit human preference supervision and provides a broader signal of perceived image quality and prompt compatibility.

\paragraph{PickScore (\texttt{pickscore}).}
PickScore~\citep{kirstain2023pick} is a CLIP-based preference model trained on pairwise human feedback from Pick-a-Pic. It assigns a scalar score to a prompt--image pair using the scaled similarity between its learned text and image embeddings. Training encourages the relative scores of competing images to match observed human choices. Like HPS, PickScore provides a population-level preference signal; the two models are trained on different preference datasets and may therefore rank the same images differently.

\paragraph{Combining reward dimensions.}
Because the five metrics have different numerical scales, we standardize each dimension over the sampled images before constructing synthetic preferences, as described in Section 4.1. Let $\widetilde{r}_j(c,x)$ denote the standardized score of dimension $j$. For a synthetic user with ground-truth weights $\lambda^\star$, the reward is

$$
r_{\lambda^\star}(c,x)
=
\sum_{j=1}^{5}\lambda_j^\star \widetilde{r}_j(c,x).
$$

For each image pair associated with the same prompt, we assign the synthetic preference label to the image with the higher weighted reward. This construction allows us to evaluate preference learning against a known reward function.

\subsection{Pick-a-Pic Dataset}

Pick-a-Pic~\citep{kirstain2023pick} is an open dataset of real user preferences for text-to-image generation, collected through a web application in which users submit prompts and compare generated images. Each comparison associates a text prompt with two candidate images and the user's preference feedback. The dataset also permits ties and includes user identifiers, enabling preference observations to be grouped into individual user histories. Its prompts and judgments arise from users interacting with image generation systems, providing a setting for studying heterogeneous visual preferences.

We use Pick-a-Pic v1 in two experimental settings. In the controlled toy experiments, it supplies captioned image pairs, while preference labels are constructed from the synthetic weighted reward rather than the original human choices. Each repetition samples 200 pairs, with 100 used for fitting the preference weights and 100 for evaluation. In the real-user experiments, we use the recorded human choices to fit and evaluate personalized rewards. Following Section 4.2, we select the 631 users with the largest numbers of preference pairs from the training split, hold out 50 pairs per user for evaluation, and use the remaining pairs for training. These experiments assess, respectively, recovery of controlled synthetic preferences and prediction of unseen choices from real users.

\subsection{Toy Example $\lambda$ Weights}\label{app:toy-exp-weights}

Table~\ref{tab:toy-configurations} reports the ground-truth and fitted preference weights for the five synthetic users introduced in Section~\ref{sec:toy-exp-sec}, with fitted weights averaged over three independently sampled datasets. The configurations cover both broadly distributed preferences (uniform and half-normal) and more concentrated preferences (Pareto). Weight recovery is strongest for the heavy-1 configuration, with a reported cosine similarity of 0.990, and the fitted vector preserves its emphasis on the third and fourth reward dimensions. Recovery is less accurate for the uniform and half-normal configurations, where the fitted weights place greater emphasis on the fifth dimension. These results highlight that recovering the exact preference weights remains challenging with limited pairwise feedback, even in this controlled setting.

\begin{table}[h]
    \centering
    \caption{Ground-truth and fitted preference weights for five synthetic
    users with uniform, half-normal, or Pareto weight configurations.
    Each row reports the ground-truth vector $\lambda^\star$, the fitted
    vector $\hat\lambda$ averaged over three datasets, and the reported cosine  similarity between fitted and ground-truth weights (higher is better). }
    \label{tab:toy-configurations}
    \small
    \setlength{\tabcolsep}{3pt}
    \begin{tabular}{l l c c c}
        \toprule
        Config & Family & \shortstack{Ground-truth\\$\lambda^\star$} & \shortstack{Fitted $\hat\lambda$\\(mean over 3 datasets)} & $\cos(\hat\lambda,\lambda^\star)$ \\
        \midrule
        uniform & 5-dim uniform & $(0.20,0.20,0.20,0.20,0.20)$ & $(0.12,0.09,0.16,0.22,0.41)$ & 0.869 \\
        normal-1 & half-normal & $(0.22,0.17,0.19,0.18,0.24)$ & $(0.15,0.06,0.13,0.19,0.47)$ & 0.870 \\
        normal-2 & half-normal & $(0.24,0.11,0.18,0.25,0.22)$ & $(0.17,0.00,0.12,0.27,0.44)$ & 0.883 \\
        heavy-1 & Pareto & $(0.06,0.03,0.44,0.40,0.07)$ & $(0.00,0.00,0.51,0.39,0.10)$ & 0.990 \\
        heavy-2 & Pareto & $(0.38,0.06,0.10,0.26,0.20)$ & $(0.34,0.00,0.04,0.25,0.37)$ & 0.939 \\
        \bottomrule
    \end{tabular}
\end{table}

\subsection{Ablation on Regularization}\label{app:lambda-reg}

\begin{figure}[htbp]
    \centering
    \begin{minipage}[h]{0.48\textwidth}
        \vspace{0pt}
        \centering
        \includegraphics[width=\linewidth]{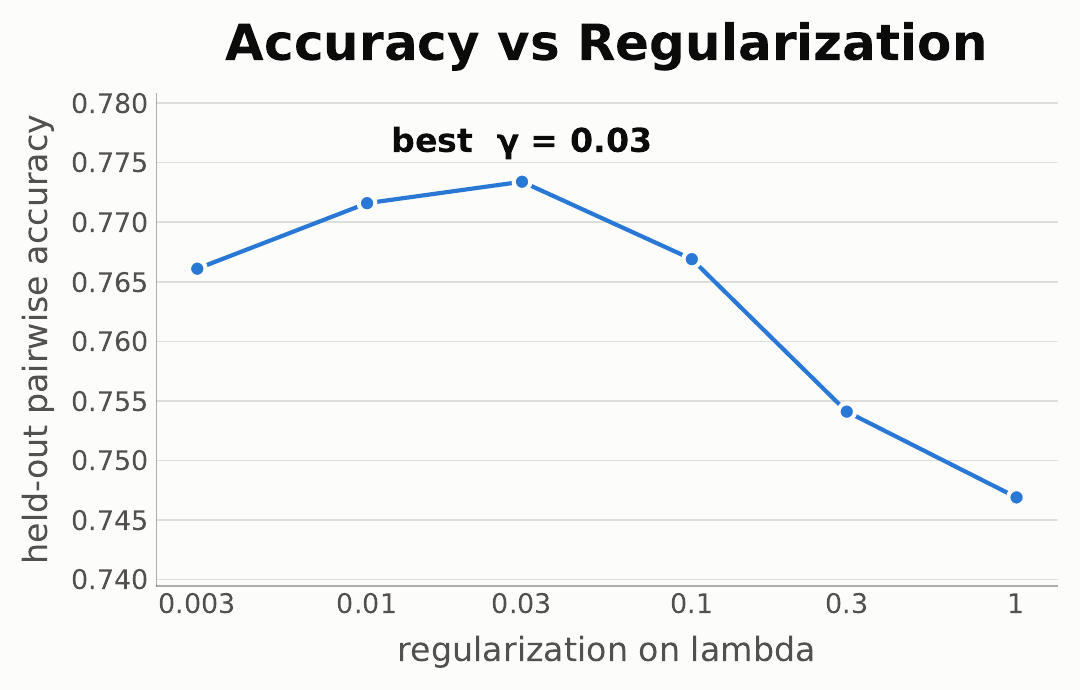}
        \caption{Ablation on the regularization coefficient $\gamma$.}
        \label{fig:gamma-sweep}
    \end{minipage}\hfill
    \begin{minipage}[h]{0.48\textwidth}
        \vspace{0pt}
        \begingroup
    \centering
    \captionof{table}{Comparison of correlation-based dimension reduction, PCA based dimension reduction, and an  autoencoder across four preference-space dimensions $k$. Values are reported as percentages. }
    \label{tab:dimension-reduction}
    \setlength{\tabcolsep}{4pt}
    \begin{tabular}{c c c c }
        \toprule
        $k$ & Correlation & PCA & Autoencoder  \\
        \midrule
        20 & 74.52\% & 74.41\% & 75.28\% \\
        30 & 74.81\% & 75.61\% & 76.29\% \\
        40 & 74.87\% & 76.05\% & 76.88\% \\
        50 & 75.39\% & 76.63\% & 77.14\% \\
        \bottomrule
    \end{tabular}
\endgroup
    \end{minipage}
\end{figure}

Figure~\ref{fig:gamma-sweep} examines the effect of the regularization coefficient $\gamma$ on held-out pairwise preference accuracy. Among the tested values, accuracy peaks at approximately 77.3\% with $\gamma=0.03$. Increasing $\gamma$ beyond this value progressively reduces accuracy, reaching approximately 74.7\% at $\gamma=1$. This pattern suggests that moderate regularization provides a useful balance between fitting observed preferences and generalizing to unseen comparisons, whereas excessive regularization limits the model's ability to capture user-specific preferences.

\subsection{Dimension Reduction with Correlation}\label{app:corr-ablation}

\paragraph{Correlation-based dimension construction.}
Let $a(x)\in{0,1}^{512}$ denote the attribute annotation vector for image $x$. We first compute pairwise correlations between attribute dimensions across the annotated images, then group highly correlated attributes into $k$ clusters, denoted by $\mathcal{C}_1,\ldots,\mathcal{C}_k$. Each cluster defines one shared preference dimension. For an image $x$, we compute its score along dimension $j$ by averaging the binary labels of the attributes belonging to that cluster:

$$
z_j^{\mathrm{corr}}(x)
=
\frac{1}{|\mathcal{C}_j|}
\sum_{\ell\in\mathcal{C}_j} a_\ell(x),
\qquad j=1,\ldots,k.
$$

For example, if a cluster contains ten attributes and four are present in an image, its score is $0.4$. This aggregation converts the original binary attribute vector into a compact representation $z^{\mathrm{corr}}(x)\in[0,1]^k$, where each coordinate measures the fraction of attributes present within a correlated group.

\paragraph{Principal component analysis.}
We also consider PCA as a linear alternative to the autoencoder for constructing the shared attribute representation. We use the same annotated image pool, with each image represented by its binary attribute vector $a(x)\in\{0,1\}^{512}$. Let $\bar{a}$ denote the mean attribute vector over this pool. We center the attribute vectors and compute the $k$ principal directions with the largest eigenvalues of the empirical covariance matrix. Collecting these directions as the columns of $W_k\in\mathbb{R}^{512\times k}$, we represent each image as
\begin{equation}
    z^{\mathrm{PCA}}(x)
    =
    W_k^{\top}\bigl(a(x)-\bar{a}\bigr)
    \in\mathbb{R}^{k}.
\end{equation}
The mean vector and projection matrix are estimated from the representation-learning pool and subsequently held fixed. Unlike correlation-based grouping, which averages attributes within each cluster, PCA constructs orthogonal linear combinations that retain the greatest attribute variance at the chosen dimensionality. In contrast to the autoencoder, its representation is restricted to a linear projection.

\paragraph{Evaluator and personalized reward.}
For both alternatives, we replace the autoencoder latent targets with the corresponding $k$-dimensional representation while retaining the evaluator architecture and training objectives described in Appendix~\ref{app:evaluator-train}. Specifically, we compute the representation difference between the two images in each training pair. For each dimension, we select the top 33\% of pairs by absolute difference and use the sign of the difference as the binary attribute-ordering target. The evaluator is trained using this dimension-specific supervision together with human preference supervision. We then fit user-specific weights under the same personalized reward-learning procedure:
\begin{equation}
    r_u(c,x)=\lambda_u^{\top}s_{\theta}(c,x).
\end{equation}
This comparison varies the construction of the shared attribute representation while keeping the downstream evaluator and personalization procedure unchanged.

\section{Prompts for Profile-Aware Evaluation}
\label{app:prompts}

Table~\ref{tab:judge-prompt-templates} presents the two-stage prompting protocol used for profile-aware evaluation. First, GPT-5.5 constructs a user preference profile from 10 image pairs, one for each test prompt. Within each pair, the preferred image is followed by the dispreferred image, allowing the model to summarize both favored and rejected visual characteristics. The profile covers subject matter and stylistic preferences while explicitly excluding judgments of image quality or rendering fidelity. Second, GPT-5.5 receives this profile, an evaluation caption, and two generated images, and selects the image that better matches the user's preferences. In the templates, \texttt{\{k\}} and \texttt{\{n\}} are set to 10 and 20, respectively; \texttt{\{profile\}} and \texttt{\{caption\}} are replaced with the generated profile and evaluation caption.

\begin{table}[t]
\centering
\caption{\textbf{Prompt templates for user profiling and GPT-5.5 evaluation.}
The first prompt constructs a taste profile from preferred--dispreferred image pairs.
The second compares two generated images using that profile.
Here, \texttt{\{k\}} is the number of preference pairs and \texttt{\{n\}} is the number of attached images ($n=2k$); \texttt{\{profile\}} and \texttt{\{caption\}} denote the resulting profile and evaluation caption.}
\label{tab:judge-prompt-templates}
\begingroup
\small
\setlength{\tabcolsep}{7pt}
\renewcommand{\arraystretch}{1.12}
\begin{tabularx}{\linewidth}{
    @{}>{\raggedright\arraybackslash}p{0.15\linewidth}
    >{\raggedright\arraybackslash}X@{}
}
\toprule
\rowcolor{gray!10}
\textbf{Stage} & \textbf{Prompt template} \\
\midrule
\textbf{1. User profiling}
&
You are profiling one user's personal visual taste from \texttt{\{k\}} preference examples.
\par\medskip
\texttt{\{n\}} images are attached in \textbf{THIS EXACT ORDER}, as \texttt{\{k\}} consecutive pairs:
\par\smallskip
image 1 = \textbf{PREFERRED}, image 2 = \textbf{DISPREFERRED} (same caption)
\par\smallskip
image 3 = \textbf{PREFERRED}, image 4 = \textbf{DISPREFERRED} (same caption)
\par\smallskip
... and so on for all \texttt{\{k\}} pairs.
\par\medskip
Within each pair the two images answer the same prompt, so what separates them is taste, not subject matter. Study what the preferred image has that the dispreferred one lacks, and what the dispreferred one has that the user rejected.
\par\medskip
Write a concise profile of this user's taste, in two short paragraphs:
\par\medskip
1. \textbf{TOPICS/subjects} they are drawn to (what the images are of).
\par\smallskip
2. Their \textbf{VISUAL/AESTHETIC preferences}, described using these aspects where relevant: Art Style, Brushstrokes, Color Palette, Composition, Skill, Detail, Hues, Saturation, Line, Shape, Value, Pattern, Texture, Artistic Medium, Mood, Perspective, Depth, Movement, Form, Juxtaposition, Iconography. State what they favour \textbf{AND} what they consistently reject.
\par\medskip
\textbf{IMPORTANT:} describe only \textbf{CONTENT} topics and \textbf{STYLISTIC} preferences. Do \textbf{NOT} mention image quality, resolution, sharpness, realism-as-quality, rendering fidelity, artifacts, or which images are 'better made' --- those are not part of the user's taste. Output only the profile text.
\\
\midrule
\textbf{2. Pairwise judging}
&
Here is a profile of a user's visual taste (topics + stylistic preferences):
\par\medskip
\texttt{\{profile\}}
\par\medskip
Two new AI-generated images are attached: \texttt{img\_P} (first) and \texttt{img\_Q} (second), both for the caption ``\texttt{\{caption\}}''. Based \textbf{ONLY} on how well each image matches this user's \textbf{TOPICS} and \textbf{STYLISTIC} preferences above --- \textbf{NOT} on raw image quality or which looks better made --- which image would this user more likely prefer? Answer with ``\texttt{P}'' or ``\texttt{Q}''.
\\
\bottomrule
\end{tabularx}
\endgroup
\end{table}

\section{Image Generation with Direct Noise Optimization}
\label{app:dno}

\paragraph{Deterministic sampling process.}
We apply Direct Noise Optimization (DNO)~\citep{tang2024tuning} to maximize
the learned personalized reward at inference time while keeping the image
generator frozen. Stable Diffusion 3.5 uses a flow-matching model with a
deterministic ODE-based sampler in our implementation. Once the prompt,
initial latent, and sampling configuration are fixed, no additional
randomness is introduced during denoising. Consequently, the complete
generation process can be represented as a deterministic mapping
\begin{equation}
    x = M_{\phi}(c,x_T),
\end{equation}
where $c$ is the text prompt, $x$ is the clean image, $x_T\in\mathbb{R}^{d}$ is the initial latent noise, and $\phi$ denotes the frozen diffusion generator, including its denoising sampler and VAE decoder. The initial latent is sampled from the Gaussian prior, $x_T^{(0)}\sim\mathcal{N}(0,I)$. This setting differs from the main stochastic formulation studied by \citet{tang2024tuning}, which jointly optimizes the initial noise and the additional noises injected at individual sampling steps. Because our deterministic sampler introduces no step-wise noise, its output is fully determined by $x_T$, and the initial latent is the only optimization variable.

\paragraph{Noise optimization.}
For a user $u$ with personalized reward function $r_u(c,x)$, the unregularized DNO objective is
\begin{equation}
    \max_{x_T}
    \quad
    r_u\!\left(c,M_{\phi}(c,x_T)\right).
\end{equation}
At each optimization step, we run the complete differentiable sampling trajectory from the current initial latent and evaluate the personalized reward on the resulting image. Gradients are then backpropagated through the frozen generator to $x_T$. Although gradients pass through $M_{\phi}$ and $r_u$, all parameters of the generator and personalized reward model remain fixed; only $x_T$ is updated. Starting from $x_T^{(0)}\sim\mathcal{N}(0,I)$, a gradient-ascent update can be written as
\begin{equation}
    x_T^{(k+1)}
    =
    x_T^{(k)}
    +
    \ell
    \left.
    \nabla_{x_T}
    r_u\!\left(c,M_{\phi}(c,x_T)\right)
    \right|_{x_T=x_T^{(k)}},
    \label{eq:dno-update-unregularized}
\end{equation}
where $\ell$ denotes the optimization step size. In practice, we perform this update using Adam for 30 optimization steps with learning rate 0.005.

\section{More Qualitative Results}\label{app:more-qualitative}

\begin{figure*}[h]
\centering
\setlength{\tabcolsep}{1.5pt} \begin{tabular}{ p{0.25\textwidth}  p{0.18\textwidth}  p{0.18\textwidth} p{0.18\textwidth} p{0.18\textwidth}}\toprule
\multicolumn{1}{c}{} & \multicolumn{1}{c}{} & \multicolumn{1}{c}{} & \multicolumn{1}{c}{} & \multicolumn{1}{c}{} \\

\multicolumn{1}{c}{\multirow{-2}{*}{\textbf{User Profile}}} &  
\multicolumn{1}{c}{\multirow{-2}{*}{\textbf{SD 3.5}}} & \multicolumn{1}{c}{\multirow{-2}{*}{\textbf{SD 3.5+Profile}}} & \multicolumn{1}{c}{\multirow{-2}{*}{\textbf{PPD}}} &  \multicolumn{1}{c}{\multirow{-2}{*}{\textbf{PAT}}} \\

\midrule
 
\multirow[t]{2}{0.25\textwidth}{\raggedright\footnotesize
They favor \textcolor{Green}{cooler or earthy palettes with selective strong accents, textured landscapes}... They consistently reject \textcolor{red}{overly polished fantasy spectacle, high-saturation blue/green casts}} &
\includegraphics[scale=0.135,valign=t]{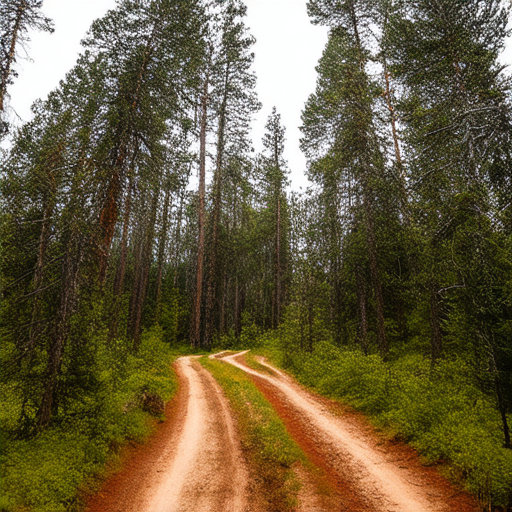} \centering
&  \includegraphics[scale=0.135,valign=t]{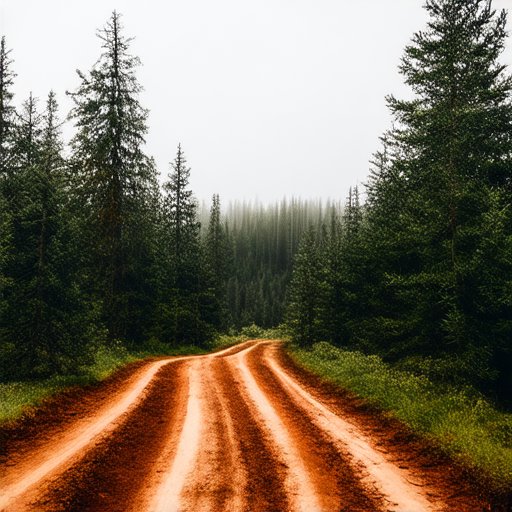} \centering
& \includegraphics[scale=0.135,valign=t]{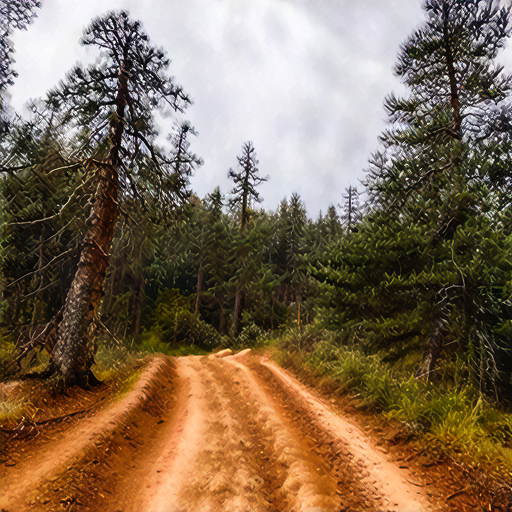} \centering
& \includegraphics[scale=0.135,valign=t]{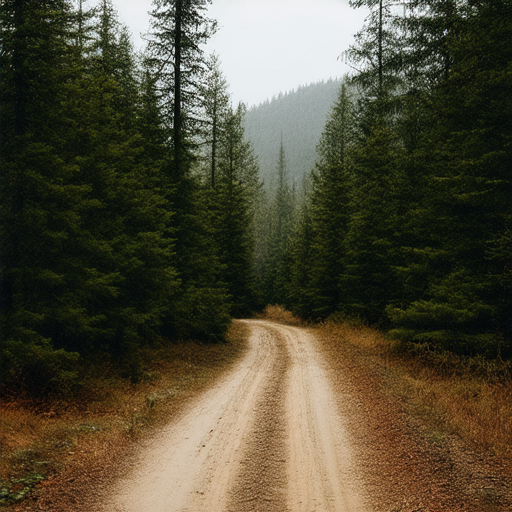} \\
& \multicolumn{4}{c}{\emph{\small photo of a dirt road in a forest, pine trees}}  \\ \midrule

\multirow[t]{2}{0.25\textwidth}{\footnotesize 
Visually, they favor clean, polished, \textcolor{Green}{high-saturation images with bright eyes, direct engagement, centered compositions, warm or vivid palettes}, smooth surfaces, and clear character silhouettes.}
&  \includegraphics[scale=0.135,valign=t]{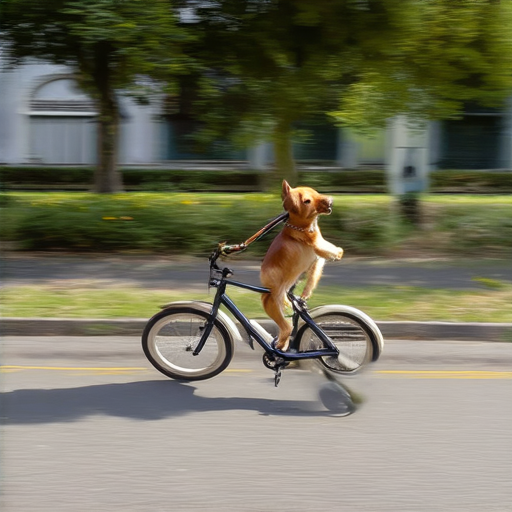} 
&  \includegraphics[scale=0.135,valign=t]{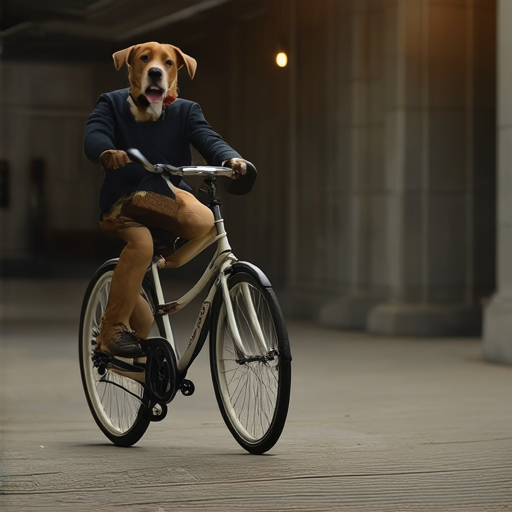}  
& \includegraphics[scale=0.135,valign=t]{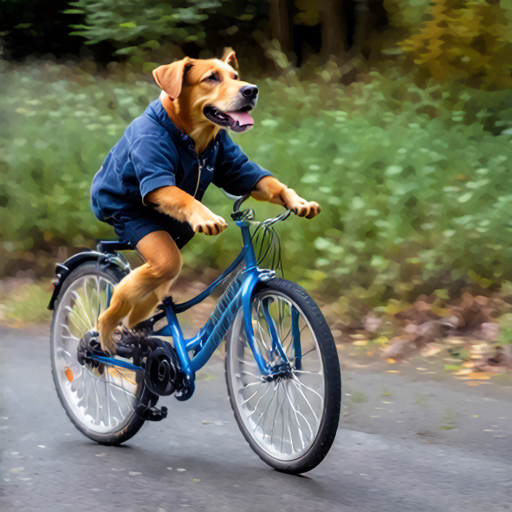} 
& \includegraphics[scale=0.135,valign=t]{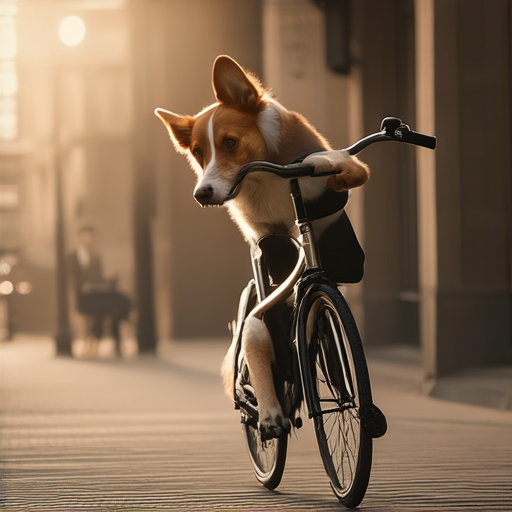} \\
& \multicolumn{4}{c}{\emph{\small a dog riding bicycle}}  \\ \midrule

\multirow[t]{2}{0.25\textwidth}{\footnotesize 
Aesthetically, they prefer vivid color, warm natural light, .... They favor \textcolor{Green}{recognizable forms with emotional immediacy}, saturated blues / oranges, and compositions centered on a single living figure. }
&  \includegraphics[scale=0.135,valign=t]{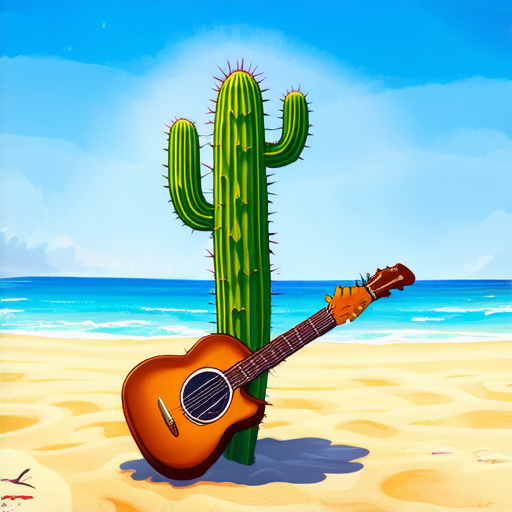} 
&  \includegraphics[scale=0.135,valign=t]{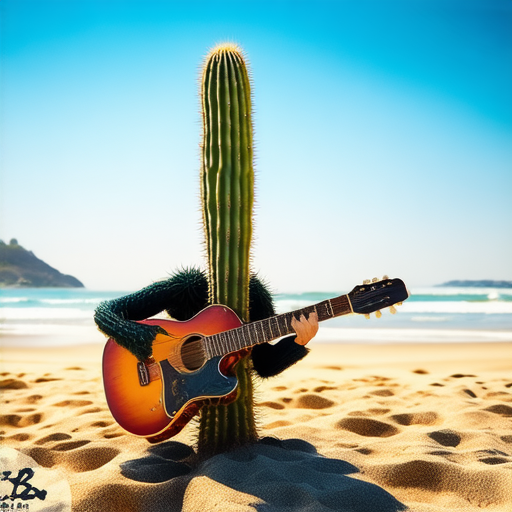}  
& \includegraphics[scale=0.135,valign=t]{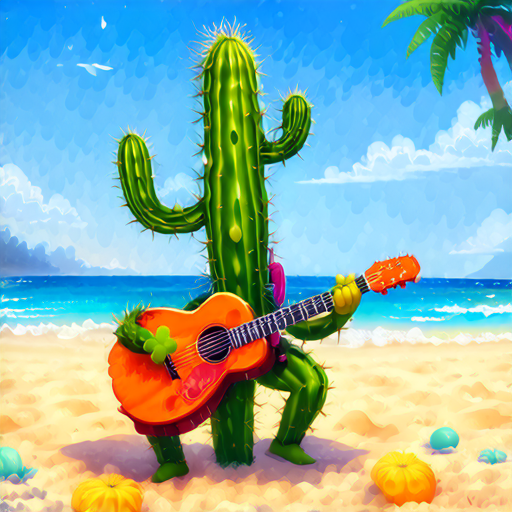} 
& \includegraphics[scale=0.135,valign=t]{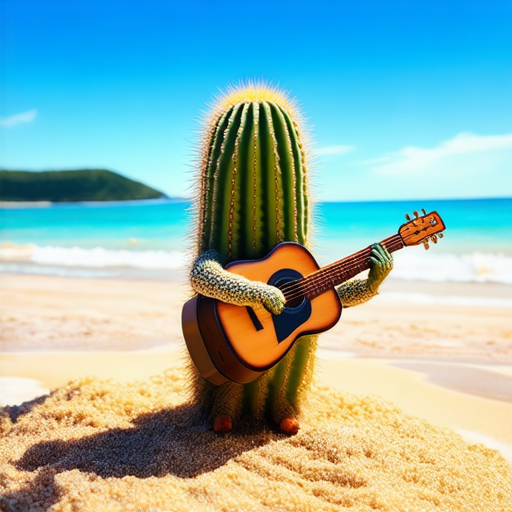} \\
& \multicolumn{4}{c}{\emph{\small A cactus playing a guitar on a beach}}  \\ \midrule
\multirow[t]{2}{0.25\textwidth}{\footnotesize 
Visually, they favor ornate, composed images with strong framing, clear central forms, full-scene context, \textcolor{Green}{rich pattern, decorative linework, and a sense of depth}.}
 &  \includegraphics[scale=0.135,valign=t]{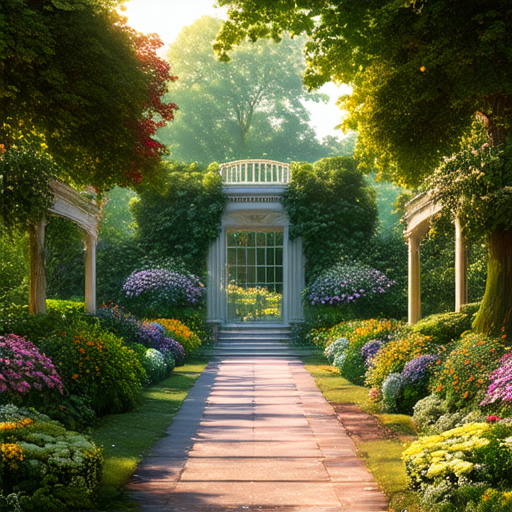} 
&  \includegraphics[scale=0.135,valign=t]{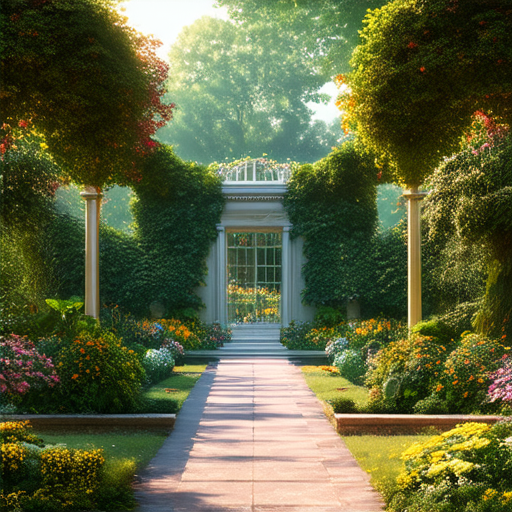}  
& \includegraphics[scale=0.135,valign=t]{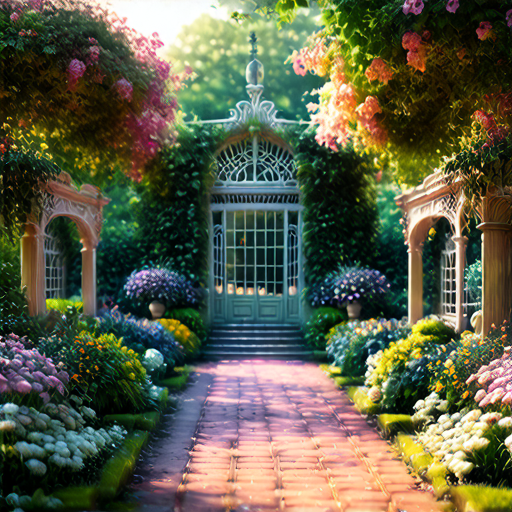} 
& \includegraphics[scale=0.135,valign=t]{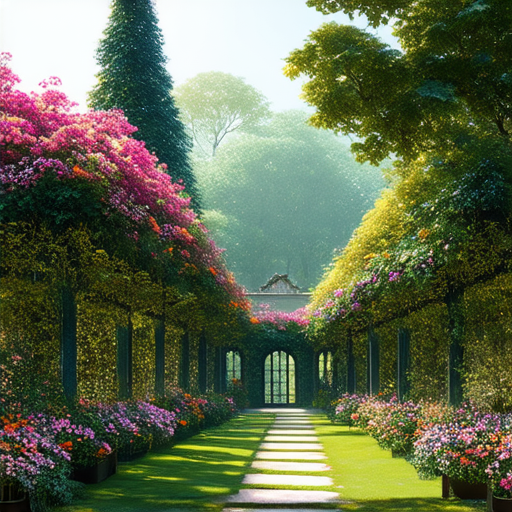} \\
& \multicolumn{4}{c}{\emph{\small concept art of a beautiful english garden, 4k, vivid colors, intricate details}}  \\ 
& \multicolumn{4}{c}{\emph{\small very realistiv, cinematic lighting, volumetric lighting}}  \\ \midrule
\multirow[t]{2}{0.25\textwidth}{\footnotesize 
Aesthetically, they favor \textcolor{Green}{vivid, strong value contrast}, glowing bokeh or star patterns, and theatrical central compositions... reject \textcolor{red}{latter, quieter, greyer, more literal scenes}; sparse compositions..}
 &  \includegraphics[scale=0.135,valign=t]{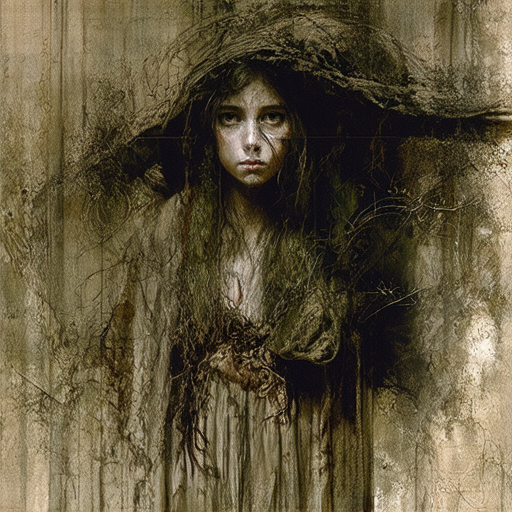} 
&  \includegraphics[scale=0.135,valign=t]{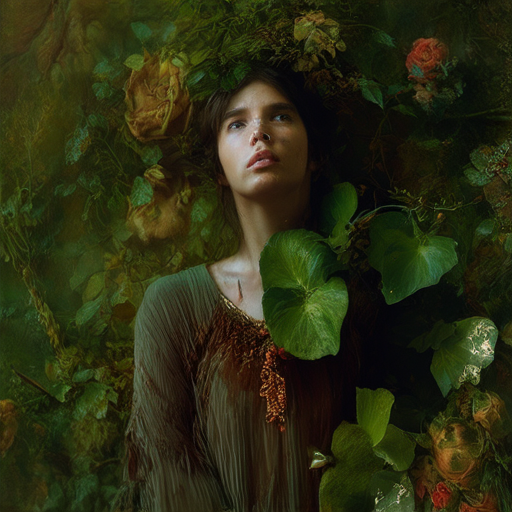}  
& \includegraphics[scale=0.135,valign=t]{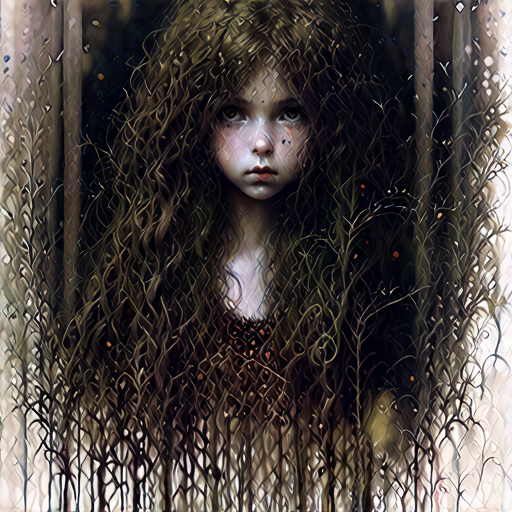} 
& \includegraphics[scale=0.135,valign=t]{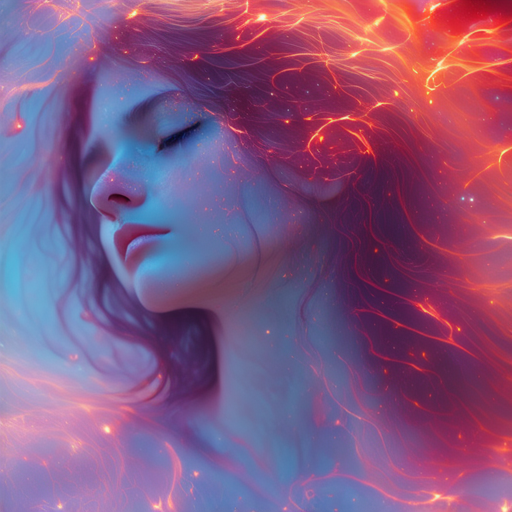} \\
& \multicolumn{4}{c}{\emph{\small sorrow}}  \\ 
\midrule
\multirow[t]{2}{0.25\textwidth}{\footnotesize 
They favor grounded compositions with environmental context, gentle depth, \textcolor{Green}{tactile textures, and stylized anime/painterly} looks when the subject is fantasy. }
 &  \includegraphics[scale=0.135,valign=t]{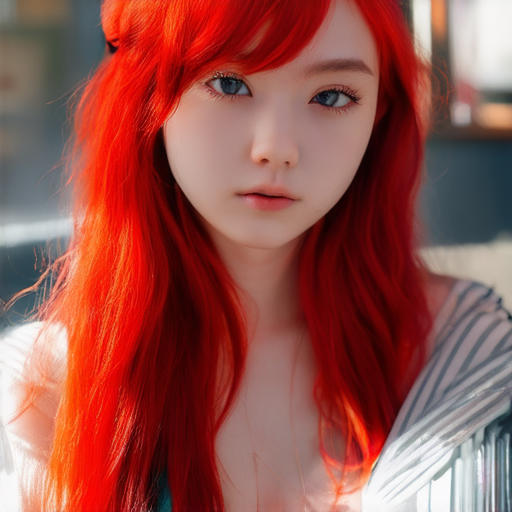} 
&  \includegraphics[scale=0.135,valign=t]{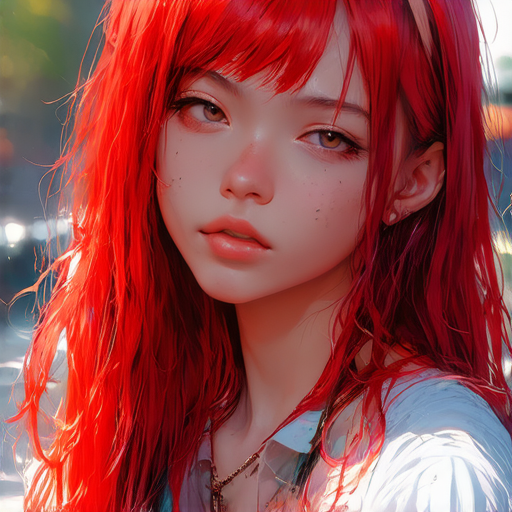}  
& \includegraphics[scale=0.135,valign=t]{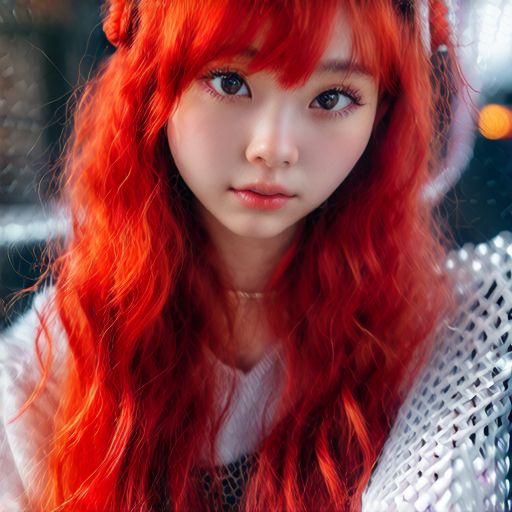} 
& \includegraphics[scale=0.135,valign=t]{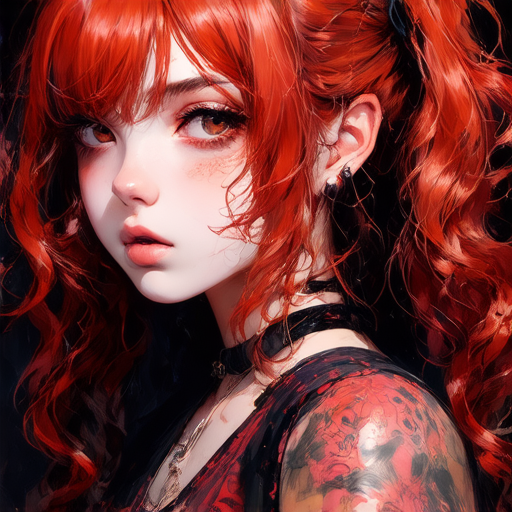} \\
& \multicolumn{4}{c}{\emph{\small Egirl with red hair, gorgeous, high-quality, beautiful}}  \\ 
\midrule
\multirow[t]{2}{0.25\textwidth}{\footnotesize 
Aesthetically, they favor saturated color, \textcolor{Green}{strong lighting, warm highlights, rich contrast}, .... They reject flatter, \textcolor{red}{emptier, more muted or gray compositions} .... }
 &  \includegraphics[scale=0.135,valign=t]{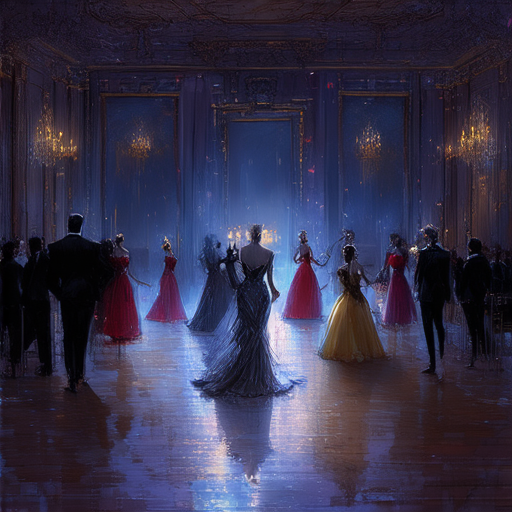} 
&  \includegraphics[scale=0.135,valign=t]{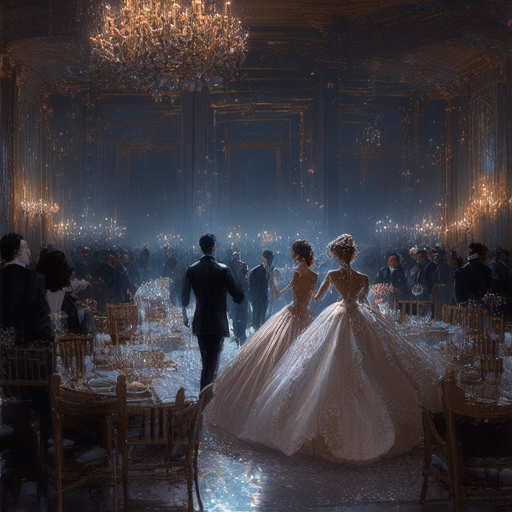}  
& \includegraphics[scale=0.135,valign=t]{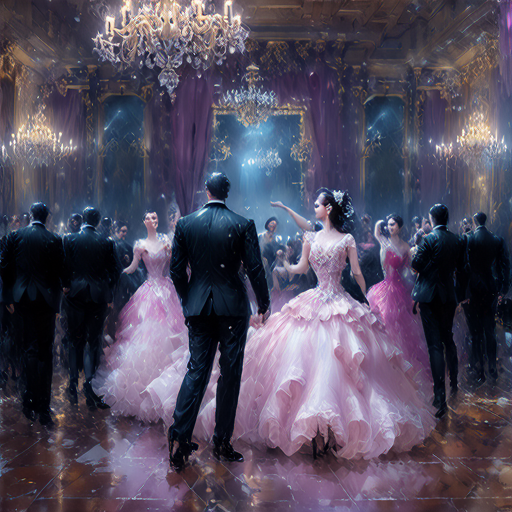} 
& \includegraphics[scale=0.135,valign=t]{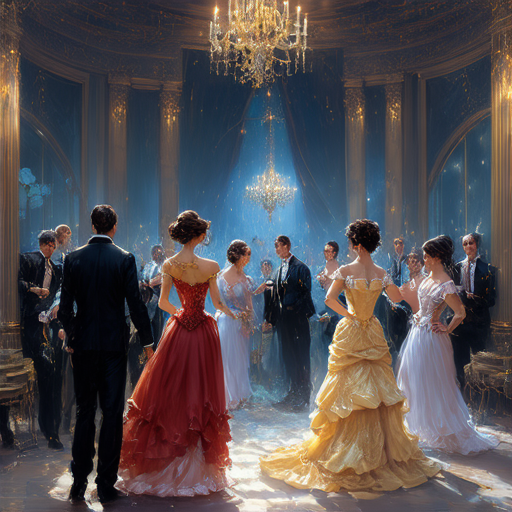} \\
& \multicolumn{4}{c}{\emph{\small concept art, drawing of fantasy ballroom at night with faceless dancers}}  \\ 
\bottomrule
    \end{tabular}
    \caption{More qualitative comparison of personalization methods.}
    \label{fig:user_profile_qualitative_more}
\end{figure*}

We add more qualitative examples in Figure~\ref{fig:user_profile_qualitative_more}. In general, \ourmech{} generates images that are closer to user profile.  

\end{document}